\documentclass[11pt]{article}
\usepackage[preprint]{acl}

\usepackage{times}
\usepackage{latexsym}
\usepackage[T1]{fontenc}
\usepackage[utf8]{inputenc}
\usepackage{microtype}
\usepackage{inconsolata}
\usepackage{graphicx}

\usepackage{amsmath}
\usepackage{dsfont}
\usepackage{subcaption}
\usepackage{todonotes}
\usepackage{listings}
\usepackage[dvipsnames]{xcolor}
\usepackage{booktabs}

\definecolor{bgblue}{rgb}{0.93, 0.96, 1}
\definecolor{borderblue}{rgb}{0.0, 0.24, 0.53}

\lstdefinestyle{llmprompt}{
    basicstyle=\ttfamily\tiny,
    breaklines=true,
    frame=single,
    rulecolor=\color{borderblue},
    backgroundcolor=\color{bgblue},
}

\title{Comprehensiveness Metrics for Automatic Evaluation\\of Factual Recall in Text Generation}

\author{
	Adam Dejl\textsuperscript{1,}\thanks{Work done while at IBM Research}, James Barry\textsuperscript{2}, Alessandra Pascale\textsuperscript{2} and Javier Carnerero Cano\textsuperscript{2}\medskip \\
	\textsuperscript{1}Department of Computing, Imperial College London \\
	\textsuperscript{2}IBM Research\medskip \\
	\textbf{Correspondence:} adam.dejl18@imperial.ac.uk, javier.cano@ibm.com
}

\begin{document}
\maketitle
\begin{abstract}
Despite demonstrating remarkable performance across a wide range of tasks, large language models (LLMs) have also been found to frequently produce outputs that are incomplete or selectively omit key information. In sensitive domains, such omissions can result in significant harm comparable to that posed by factual inaccuracies, including hallucinations. In this study, we address the challenge of evaluating the comprehensiveness of LLM-generated texts, focusing on the detection of missing information or underrepresented viewpoints. We investigate three automated evaluation metrics: (1) an \emph{NLI-based} method that decomposes texts into atomic statements and uses natural language inference (NLI) to identify missing facts, (2) a \emph{Q\&A-based} metric that extracts question-answer pairs and compares responses across sources, and (3) an \emph{end-to-end} approach that directly identifies missing content using LLMs. Our experiments demonstrate the surprising effectiveness of the simple end-to-end metric compared to more complex metrics, though at the cost of reduced robustness, interpretability and result granularity. We further assess the comprehensiveness of responses from several popular open-weight LLMs when answering user queries based on multiple sources.
\end{abstract}

\section{Introduction}
\label{sec:intro}

\begin{figure}[!ht]
    \centering
    \begin{subfigure}[m]{0.99\linewidth}
        \centering
        \includegraphics[width=0.99\linewidth]{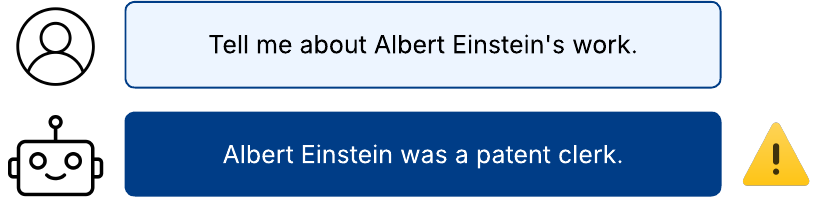}
        \caption{Example of an LLM response omitting key details.}
        \label{fig:example-einstein-clerk}
    \end{subfigure}
    \par\medskip
    \begin{subfigure}[m]{0.99\linewidth}
        \centering
        \includegraphics[width=0.99\linewidth]{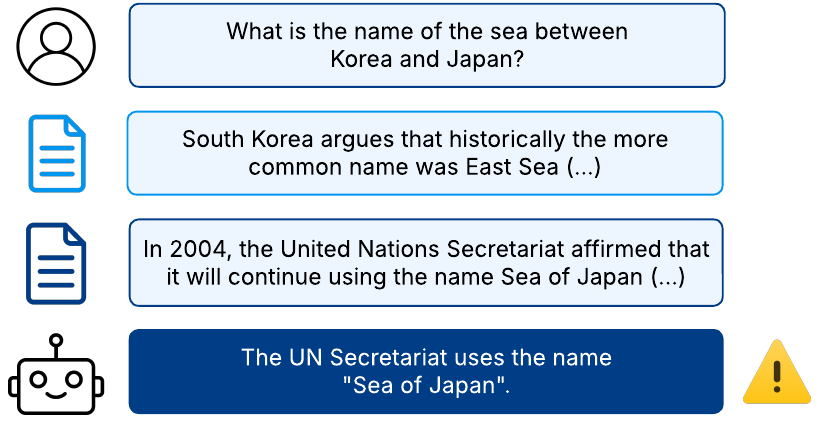}
        \caption{Example of an LLM response selectively favouring a specific viewpoint.}
        \label{fig:example-sea-name}
    \end{subfigure}
    \caption{Illustration of two scenarios in which comprehensiveness evaluation could reveal issues with the model responses\protect\footnotemark. \textbf{(a)} The model provides a factually precise but incomplete answer, failing to mention more important and relevant facts. \textbf{(b)} Despite conflicting evidence, the LLM only presents a one-sided view without acknowledging the conflict.}
    \label{fig:comprehensiveness-examples}
\end{figure}

Application of large language models (LLMs) in safety-critical domains often hinges on their ability to dependably produce factually correct outputs \cite{wang-etal-2024-factuality}. While significant attention has been drawn to the issue of hallucinations, where the model produces outputs inconsistent with the facts, omissions of key information or misrepresentation of conflicting evidence can result in similarly severe consequences \cite{busch-2025-llm-applications-care}. There is thus a clear need for automated evaluation metrics that can identify such omissions and assess factual recall in general.

\footnotetext{Icons from Noun Project (\href{https://thenounproject.com/icon/user-7933341/}{User} by mar\_yani, \href{https://thenounproject.com/icon/bot-7035238/}{bot} by Yosua Bungaran and \href{https://thenounproject.com/icon/document-5079157/}{Document} by Alzam, \href{https://creativecommons.org/licenses/by/3.0/deed.en}{CC BY 3.0}) and Antü Plasma Suite (\href{https://commons.wikimedia.org/wiki/File:Antu_dialog-warning.svg}{Antu dialog-warning} by Fabián Alexis, \href{https://creativecommons.org/licenses/by-sa/3.0/deed.en}{CC BY-SA 3.0}).}

\begin{figure*}[tb]
    \centering
    \includegraphics[width=0.9\textwidth]{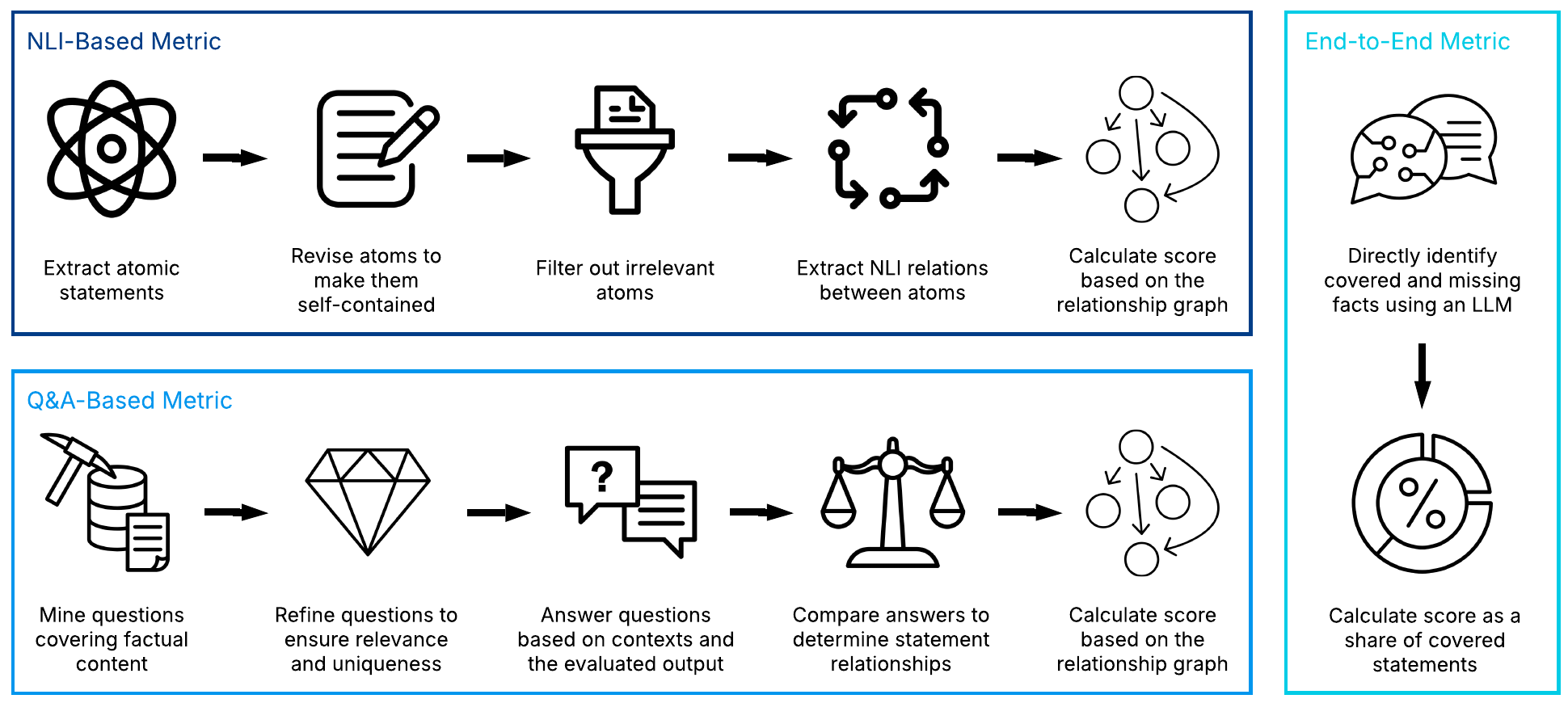}
    \caption{Overview of the three comprehensiveness metrics introduced in this work\protect\footnotemark.}
    \label{fig:comprehensiveness-versions}
\end{figure*}

Early methods for evaluating long-form factuality in LLMs have predominantly focused on factual precision while neglecting recall \cite{min-etal-2023-factscore, chern-2023-factool}. As a result, these methods cannot reliably identify incomplete model responses, as illustrated by the examples in Figure~\ref{fig:comprehensiveness-examples}. While some more recent techniques attempt to incorporate recall, they typically rely on coarse and potentially unreliable heuristics, such as the ratio of supported facts to a predefined number of factual claims that should be included in each response \citep{wei-2024-long-form-factuality, song-etal-2024-veriscore, marinescu-2025-factreasoner}. Such approaches are unable to clearly identify the relevant claims missing from the output, reducing their potential to serve as diagnostic tools or as mechanisms for real-time feedback and correction.

\footnotetext{Icons from Noun Project (\href{https://thenounproject.com/icon/atom-7938140/}{atom} by Yasashii Studio, \href{https://thenounproject.com/icon/edit-7901852/}{edit} by james art, \href{https://thenounproject.com/icon/filter-7955489/}{filter} by VERA, \href{https://thenounproject.com/icon/relation-6765983/}{relation} by Ridhia Okky Ramadhonna, \href{https://thenounproject.com/icon/graph-6170625/}{Graph} by Vectplus, \href{https://thenounproject.com/icon/data-mining-7919658/}{Data Mining} by HAZHA FARASYA, \href{https://thenounproject.com/icon/diamond-1822355/}{Diamond} by Three Six Five, \href{https://thenounproject.com/icon/questions-and-answers-7621736/}{question and answers} by Aidan Stonehouse, \href{https://thenounproject.com/icon/comparison-7633065/}{comparison} by Keyy Creative, \href{https://thenounproject.com/icon/llm-7675833/}{llm} by Gonza Monta and \href{https://thenounproject.com/icon/ratio-7661071/}{ratio} by Sulistiana, \href{https://creativecommons.org/licenses/by/3.0/deed.en}{CC BY 3.0}).}

In this work, we instead focus on directly identifying the specific pieces of information absent from a model's output. As discussed in prior literature, this task is inherently challenging since it is practically impossible to identify a definite set of atomic facts that should be included in a model response \cite{wei-2024-long-form-factuality}. To make the problem tractable, we evaluate comprehensiveness with respect to a predefined corpus of texts. This corpus is assumed to contain all key information for a specific query, though it may also include details that are irrelevant. Due to the strong capabilities of the available search engines and retrieval models that can be readily used to look up relevant texts \cite{fan-2024-rag-survey}, we believe that this assumption does not substantially reduce the applicability of our evaluation metrics.

Given a background corpus, the comprehensiveness of a model output is defined as the share of relevant atomic facts that are included in this output (see Section \ref{sec:methods} for a more formal definition). We propose three different approaches for evaluating comprehensiveness (see Figure~\ref{fig:comprehensiveness-versions}):

\begin{itemize}
    \item The \emph{NLI-based} method decomposes both the model output and corpus texts into atomic statements, filters them for relevance, constructs a relational graph, and identifies missing content via graph analysis.
    \item The \emph{Q\&A-based} method mines questions covering the factual content of the considered texts, answers them using both the model output and the corpus, and compares these answers to determine their relationships. The obtained question-answer pairs can then act as a drop-in replacement for the atomic facts used in the NLI-based strategy.
    \item The \emph{end-to-end} (E2E) method leverages LLMs to directly identify atomic pieces of information missing from the evaluated text.
\end{itemize}

To evaluate the effectiveness of these methods, we conduct experiments on the WikiContradict \cite{hou-2024-wikicontradict} and ConflictBank \cite{zhaochen-2024-conflictbank} benchmarks. These datasets consist of background texts containing conflicting pieces of information, queries eliciting this information, and responses reflecting the content of known subsets of the background texts, making them naturally suited for assessing the ability to identify incomplete answers. We also evaluate the comprehensiveness of outputs generated by several popular open-weight LLMs when applied to real-world questions from the r/explainlikeimfive (ELI5) Reddit forum. In this setting, the evaluated LLMs have to correctly synthesize information from diverse web-based sources, testing the capabilities of these models in processing factual knowledge.

In summary, the key contributions of our work are as follows:

\begin{itemize}
    \item We introduce three novel metrics for automatically evaluating the comprehensiveness of LLM outputs with respect to a reference corpus. These metrics are designed to handle the unique challenges associated with assessing factual recall, such as accounting for information relevance and considering relations between the individual facts.
    \item We assess the effectiveness of the introduced metrics (i.e., we meta-evaluate them) on the WikiContradict and ConflictBank datasets, which provide a natural test bed for identifying missing information.
    \item We evaluate comprehensiveness of several popular open-weight LLMs when answering real-world user queries based on multiple sources.
\end{itemize}

\section{Related Work}
\label{sec:related-work}

Automated evaluation of LLM factuality has been a longstanding topic of research. Standardized benchmarks such as TruthfulQA \cite{lin-etal-2022-truthfulqa}, HaluEval \cite{li-etal-2023-halueval, li-etal-2024-halueval-2}, and SimpleQA \cite{wei-2024-simpleqa, haas-2025-simpleqa-verified} assess model performance on factual question answering, flagging incorrect responses and hallucinations. Other works have focused on evaluating factual precision in long-form outputs, typically by decomposing these responses into atomic units and verifying them individually \cite{min-etal-2023-factscore, chern-2023-factool, wei-2024-long-form-factuality, song-etal-2024-veriscore, fatahi-bayat-etal-2025-factbench, marinescu-2025-factreasoner}.

A related line of research evaluates information consistency, particularly in summarization. These methods typically involve answering a shared set of questions based on different texts and comparing the answers \citep{eyal-etal-2019-question-answering, wang-etal-2020-qags, durmus-etal-2020-feqa, scialom-etal-2021-questeval, manakul-etal-2023-mqag, liu-etal-2024-summequal}.

Particularly relevant to our work are approaches aiming to assess factual recall. \citet{wei-2024-long-form-factuality} propose a simple recall metric, $R_K(y) = \min\left(\frac{S(y)}{K}, 1\right)$, where $S(y)$ is the number of verified facts in response $y$, and $K$ is a hyperparameter specifying the number of supported facts required for a full recall. However, choosing $K$ appropriate for the given use-case can be challenging, and the metric does not clearly identify the facts that are missing from the response. \citet{qi-etal-2024-long2rag} and \citet{liu-2025-verifact} assess recall by comparing model outputs to reference sets of facts. However, unlike our work, their focus is on constructing fixed benchmarks rather than general-purpose comprehensiveness evaluation of responses to arbitrary queries. \citet{samarinas-etal-2025-beyond-accuracy} evaluate information coverage in model responses, but focus on broader LLM-generated subtopics rather than specific factual claims. Finally, the AutoNuggetizer framework proposed by \citet{pradeep-2025-great-nugget-recall} automatically evaluates recall of important facts or ``nuggets.'' However, compared to our metrics, their approach is less granular and uses a fixed number of nuggets per query instead of keeping it flexible.

\section{Comprehensiveness Metrics}
\label{sec:methods}

In this section, we introduce three metrics for evaluating the comprehensiveness of LLM-generated responses. Each metric takes as input the original user prompt $P$, the evaluated model response $R$, and a corpus of contextual texts $\mathcal{C}$. Optionally, the corpus can be preprocessed via summarization to reduce computational overhead during evaluation (see Listing \ref{lst:context-summariser} for the corresponding prompt).

Given these inputs, each metric outputs two sets of atomic statements: those that are \emph{covered} by the response, denoted $\mathcal{A}_\text{in}$, and those that are \emph{uncovered} or missing, denoted $\mathcal{A}_\text{out}$. These sets are used to compute a scalar comprehensiveness score:
\begin{align*}
S = \frac{|\mathcal{A}_\text{in}|}{|\mathcal{A}_\text{in}| + |\mathcal{A}_\text{out}|}
\end{align*}
where $|.|$ denotes set cardinality. In addition to these base outputs, the NLI-based and Q\&A-based metrics provide more fine-grained results, such as graphs capturing relationships between the relevant atomic statements. We describe these specifics in more detail when introducing the individual comprehensiveness versions below.

\subsection{NLI-Based Comprehensiveness}

The NLI-based comprehensiveness metric identifies missing information in a model response by constructing a directed graph whose nodes represent atomic statements and whose edges encode entailment relations derived through natural language inference (NLI). This approach is inspired by the method previously used by \citet{marinescu-2025-factreasoner} to evaluate factual precision, although the evaluation of comprehensiveness requires additional steps such as relevance filtering and graph analysis.

\noindent \textbf{Atomic Statement Extraction.} We begin by extracting atomic statements that capture the factual content of the model response $R$ and the contextual texts in the background corpus $\mathcal{C}$. Formally:
\begin{align*}
    \forall X \in \{R\} \cup \mathcal{C} \ldotp \mathcal{A}_{\text{init}, X} := \text{AtomExtractor}(X)
\end{align*}
The AtomExtractor is implemented using an LLM with a few-shot prompt (see Listing~\ref{lst:atomic-stmt-extraction}).

\noindent \textbf{Atom Revision.} The extracted atoms are then revised to ensure they are self-contained (i.e., free of unresolved references such as ``he'' or ``the event'') and decomposed further if they contain conjunctive statements. For example, ``George Orwell wrote \textit{Animal Farm} and \textit{1984}'' would be split into two distinct claims. We define:
\begin{align*}
    \forall X \in& \{R\} \cup \mathcal{C} \ldotp \mathcal{A}_{\text{rev}, X} := \\ &\text{AtomReviser}(X, \mathcal{A}_{\text{init}, X})
\end{align*}
This step also uses an LLM with the prompt given in Listing~\ref{lst:atomic-revision}.

\noindent \textbf{Relevance Filtering.} Not all atoms extracted from $\mathcal{C}$ are relevant to the user query $P$. We filter out irrelevant statements by estimating a relevance score for each atom and discarding those below a pre-determined threshold $T_{\text{rel}}$\footnote{\label{fn:thresholds} See Appendix \ref{apd:meta-experimental-setup} for discussion and justification of our use of thresholds.}. Let:
\begin{align*}
    \mathcal{A}_R &:= \mathcal{A}_{\text{rev}, R} \\
    \forall X \in \mathcal{C} \ldotp \mathcal{A}_X &:= \{A_i \in \mathcal{A}_{\text{rev}, X} | \\
    &\phantom{:=} \text{RelEstimator}(P, \mathcal{A}_{\text{rev}, X})_i \geq T_{\text{rel}} \} \\
    \mathcal{A}_{\mathcal{C}} &:= \bigcup_{X \in \mathcal{C}} \mathcal{A}_X \quad \mathcal{A} := \mathcal{A}_R \cup \mathcal{A}_{\mathcal{C}}
\end{align*}
Relevance scores are estimated using an LLM and computed as a probability-weighted sum of the valid output scores. The prompt used for relevance filtering is given in Listing~\ref{lst:relevance-filtering}.

\noindent \textbf{NLI Relation Extraction.} Once the relevant atoms are identified, we use natural language inference (NLI) to determine the relationships between them. For comprehensiveness evaluation, we only focus on the entailment relations, though contradiction and neutral relations are also computed as by-products. Relations are extracted for all context-response, response-context and context-context atom pairs. Response-response relations are excluded, as they are irrelevant for evaluating comprehensiveness. The atoms along with the relations extracted in this step are then used to construct a directed fact graph $G_F$ as follows:
\begin{gather*}
    \mathcal{V}_F := \mathcal{A} \quad \mathcal{E}_F := \text{NliExtractor}(\mathcal{A}_R, \mathcal{A}_{\mathcal{C}}) \\
    G_F := (\mathcal{V}_F, \mathcal{E}_F)
\end{gather*}
Instead of using a dedicated NLI model, we extract the NLI relations using a standard LLM, as a general-purpose model is better able to handle complex and knowledge-heavy statements arising in comprehensiveness evaluation. The used LLM prompt is given in Listing \ref{lst:nli-relation-extraction}.

\noindent \textbf{Comprehensiveness Score Calculation.} Finally, we can determine the sets of covered and uncovered statements based on the fact graph $G_F$. Recall that $G_F$ includes atomic statements from the model response and the contextual texts as nodes and entailment relations between these statements as edges. To ensure that the statements considered in the comprehensiveness evaluation are unique, we first compute the condensation of the fact graph, $G_C = (\mathcal{V}_C, \mathcal{E}_C)$, which contracts each of its strongly connected components (SCCs) into a single node. Each SCC is represented by the mode of its constituent atoms, with precedence given to contextual atoms\footnote{This means that SCCs containing at least one contextual atom are represented by a contextual atom rather than a response atom, ensuring that all statements from the contextual texts are correctly accounted for when computing the comprehensiveness score later.}. Atoms within an SCC are treated as logically equivalent.

Let $\text{path}(A_i, A_j) \in G_F$ denote that there is a path from $A_i$ to $A_j$ in $G_F$ for some $A_i, A_j \in \mathcal{V}_F$. Then, we can determine the sets of covered and uncovered contextual statements, $\mathcal{A}_\text{in}$ and $\mathcal{A}_\text{out}$, as follows:
\begin{align*}
    \mathcal{A}_\text{in} &:= \{A_i \in \mathcal{V}_C \cap \mathcal{A}_\mathcal{C} |\\
    &\phantom{= \{.}\exists A_j \in \mathcal{A}_R\ldotp \text{path}(A_j, A_i) \in G_F\} \\
    \mathcal{A}_\text{out} &:= \{A_i \in \mathcal{V}_C \cap \mathcal{A}_\mathcal{C} | A_i \notin \mathcal{A}_\text{in}\}
\end{align*}

A contextual atom is considered covered if it is entailed by any response atom. The comprehensiveness score $S$ is then computed as described at the beginning of this section. In addition to the numerical score, we can also compute the uncovered context basis $\mathcal{A}_\text{basis} \subseteq \mathcal{A}_\text{out}$, which is the minimal set of statements that would need to be included in the response $R$ to achieve full comprehensiveness:
\begin{align*}
    \mathcal{A}_\text{basis} := \{&A_i \in \mathcal{A}_\text{out} | \\
    &\neg\exists A_j \in \mathcal{A}_\text{out}\ldotp \text{path}(A_j, A_i) \in G_F\}
\end{align*}

Intuitively, an atom $A_i$ only needs to be explicitly included if there is no more specific atom $A_j$ that entails it.

\noindent \textbf{Discussion.} While the NLI-based comprehensiveness metric provides highly fine-grained results that can be very helpful for interpreting its scores and addressing the identified issues, it also suffers from several drawbacks. In particular, it is computationally intensive, as the NLI relation extraction step requires evaluating $2 \times |\mathcal{A}_R| \times |\mathcal{A}_\mathcal{C}| + |\mathcal{A}_\mathcal{C}| \times (|\mathcal{A}_\mathcal{C}| - 1)$ relations between atoms. Moreover, the NLI extractor only considers the individual atoms without broader context when determining their relations (otherwise, this step would become even more expensive), resulting in potential misclassifications.

\subsection{Q\&A-Based Comprehensiveness}
The Q\&A-based comprehensiveness metric replaces the directly extracted atomic statements from the NLI pipeline with answers to questions about the contextual texts and the evaluated response. This approach draws inspiration from prior work on using question answering to assess information consistency \citep{eyal-etal-2019-question-answering, wang-etal-2020-qags, durmus-etal-2020-feqa, scialom-etal-2021-questeval, manakul-etal-2023-mqag, liu-etal-2024-summequal}.

\noindent \textbf{Question Mining.} In the first step of the Q\&A-based pipeline, we mine a set of open-ended questions covering the factual content of the evaluated response $R$ and the contextual texts included in $\mathcal{C}$. In particular, we extract:
\begin{align*}
    \mathcal{Q}_{\text{init}} := \bigcup_{X \in \{R\} \cup \mathcal{C}} \text{QuestionMiner}(X)
\end{align*}
Question mining is performed by an LLM instructed to extract self-contained and unambiguous questions. The used prompt is given in Listing~\ref{lst:qa-mining}.

\noindent \textbf{Question Refinement.} Since questions are extracted independently from each input text and without reference to the original user query $P$, the initial set may contain duplicates or questions that focus on irrelevant details. The refinement stage addresses this by filtering out such questions and lightly editing the remaining ones with the aim of making them more clear, general and self-contained. Similarly to the relevancy filtering in the NLI-based pipeline, we estimate a numerical relevance score for each question and discard those scoring below a threshold $T_{\text{rel}}$\textsuperscript{\ref{fn:thresholds}}:
\begin{align*}
    \mathcal{Q} := \{Q_i |& (Q_i, r_i) \in \text{QuestionRefiner}(P, \mathcal{Q}_\text{init}), \\
    &r_i \geq T_\text{rel}\}
\end{align*}
Similarly to the other stages of the pipeline, question refinement is also implemented by an LLM, with the relevance scores being probabilistically weighted according to output token likelihoods. The corresponding prompt is given in Listing \ref{lst:qa-refinement}.

\noindent \textbf{Answer Generation.} Once we have a refined set of questions, we can generate answers to these questions based on each source in $\{R\} \cup \mathcal{C}$. Each question may yield multiple or no answers per source. In addition to the answers themselves, we also elicit confidence scores indicating the degree to which each answer is considered valid by the given source and discard those below a threshold $T_{\text{conf}}$\textsuperscript{\ref{fn:thresholds}}. This ensures that low-confidence content is not flagged as missing from the evaluated response during comprehensiveness evaluation. Formally:
\begin{align*}
    \forall X &\in \{R\} \cup \mathcal{C} \ldotp \mathcal{A}_X := \{A_i | \\
        &\phantom{\in.} (A_i, c_i) \in \text{AnswerGenerator}(\mathcal{Q}, X), \\
        &\phantom{\in.} c_i \geq T_\text{conf}\} \\
    \mathcal{A}_{\mathcal{C}} &:= \bigcup_{X \in \mathcal{C}} \mathcal{A}_{X} \quad \mathcal{A} := \mathcal{A}_R \cup \mathcal{A}_{\mathcal{C}}
\end{align*}
We also define a relation $\mathcal{R} \subseteq \mathcal{Q} \times \mathcal{A}$ associating each question with its corresponding answers. Once again, we use an LLM for both generating the answers and estimating the confidence scores. The used prompt is provided in Listing \ref{lst:answer-generation}.

\noindent \textbf{Answer Comparison.} Next, we compare question answers to determine their respective relationships. We consider relation types analogous to NLI, though using a slightly different terminology more suitable for comparing pairs of answers: ``equivalent'', ``first implies second'', ``second implies first'', ``contradictory'', and ``neutral''. For consistency with the NLI-based pipeline, we convert the first three of these relation types into standard unidirectional entailments and leave the latter two types unused. Note that comparisons are only performed between answers to the same question, as comparing answers to different questions is meaningless. Given the answers as nodes and the extracted relations as edges, we obtain a fact graph $G_F$ as follows:
\begin{gather*}
    \mathcal{V}_F := \mathcal{A} \quad \mathcal{E}_F := \text{AnswerComparator}(\mathcal{Q}, \mathcal{A}, \mathcal{R}) \\
    G_F := (\mathcal{V}_F, \mathcal{E}_F)
\end{gather*}
The AnswerComparator module is implemented by an LLM augmented with tool use. The provided tool enables the model to compare physical quantities in different units using the Pint library\footnote{\url{https://pint.readthedocs.io}}, as we found the base LLM capabilities to be lacking in this respect. The prompt is given in Listing \ref{lst:answer-comparison}.

\noindent \textbf{Comprehensiveness Score Calculation.} Once the fact graph $G_F$ is constructed, we can compute the sets of covered and uncovered contextual answers, $\mathcal{A}_{\text{in}}$ and $\mathcal{A}_{\text{out}}$, and derive the comprehensiveness score $S$ and uncovered context basis $\mathcal{A}_{\text{basis}}$ analogously to the NLI-based method. The same graph condensation and path-based logic applies, with answers replacing atomic statements.

\noindent \textbf{Discussion.} The Q\&A-based metric provides comparable granularity to the NLI-based approach, with the key distinction being that the graph nodes represent question answers rather than atomic statements. The key benefits of the Q\&A approach are its ability to consider wider context when generating question answers and improved efficiency, as the answer comparison logic only needs to compare answers to matching questions rather than all pairs of atoms.

\subsection{End-to-End Comprehensiveness}

The final version of comprehensiveness relies on general LLM capabilities to directly identify the covered and missing facts when supplied with the original user query $P$, the background corpus $\mathcal{C}$, and the evaluated response $R$ in a single context:
\begin{align*}
    (\mathcal{A}_\text{in}, \mathcal{A}_\text{out}) := \text{CoverageEvaluator}(P, \mathcal{C}, R)
\end{align*}

The returned $\mathcal{A}_\text{in}$ and $\mathcal{A}_\text{out}$ sets can then be used to compute the comprehensiveness score $S$ in the same manner as for the other versions of the metric. Compared to the NLI and Q\&A approaches, this method is the most computationally efficient, as it avoids the need for intermediate steps such as atom extraction, graph construction, or pairwise relation classification. However, it also provides less granular and interpretable results. The LLM prompt used for the end-to-end comprehensiveness evaluation is provided in Listing \ref{lst:coverage-evaluator}.

\section{Experiments}
\label{sec:experiments}

We consider two types of experiments associated with comprehensiveness. First, we conduct a meta-evaluation (i.e., evaluation of the evaluator) assessing the reliability of our comprehensiveness metrics in identifying incomplete model responses. Second, we use the best-performing variants of our metrics to evaluate the comprehensiveness of several popular open-weight models when answering real-world user questions from the r/explainlikeimfive (ELI5) Reddit forum.

\begin{figure*}[tp]
    \centering
    \includegraphics[width=0.86\textwidth]{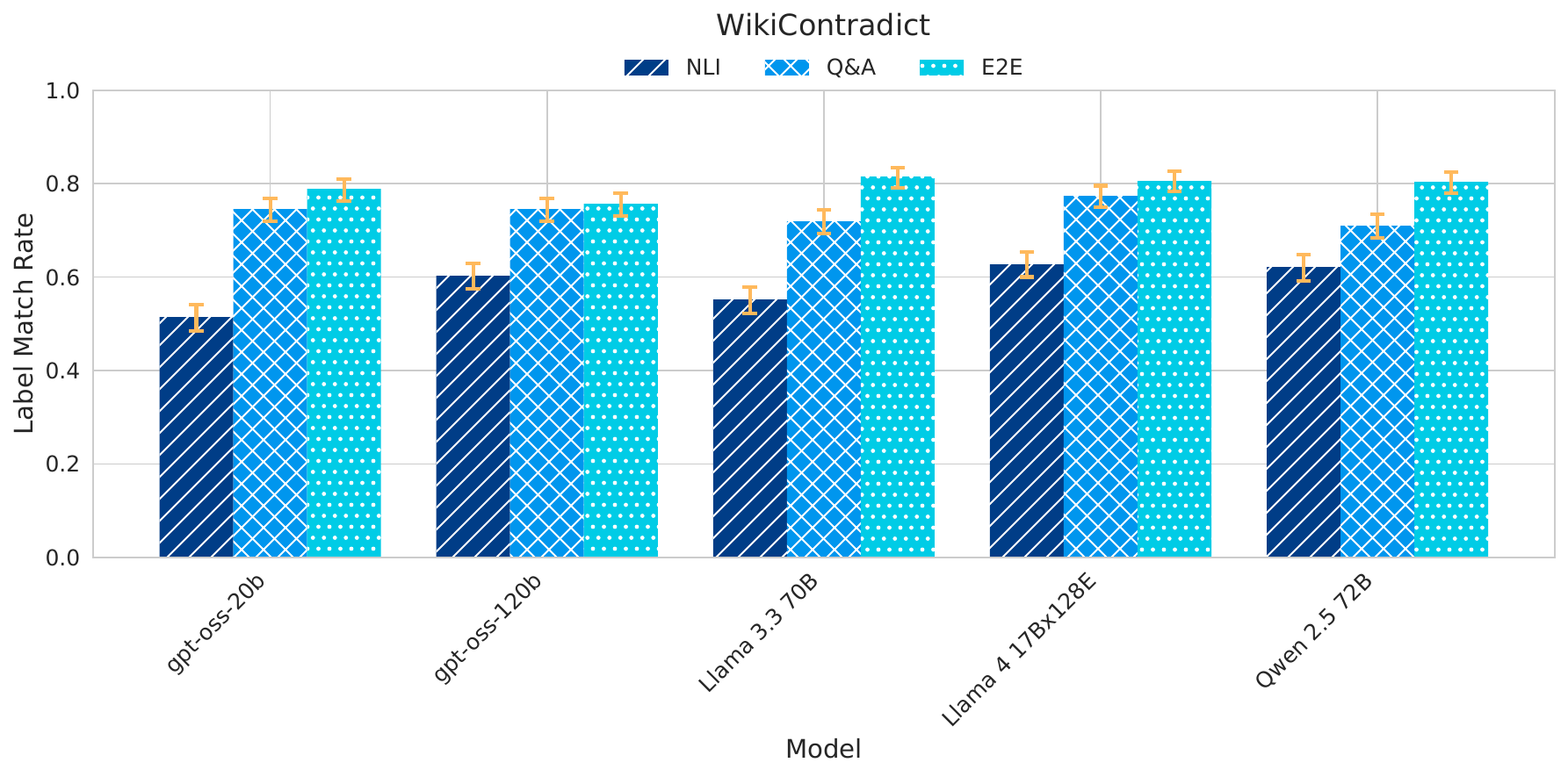}
    \caption{Results of comprehensiveness meta-evaluation on the WikiContradict dataset for different models and metric variants. The error bars indicate 95\% confidence intervals determined using BCa bootstrap.}
    \label{fig:wikicontradict-results}
\end{figure*}
\begin{figure*}[htp]
    \centering
    \includegraphics[width=0.86\textwidth]{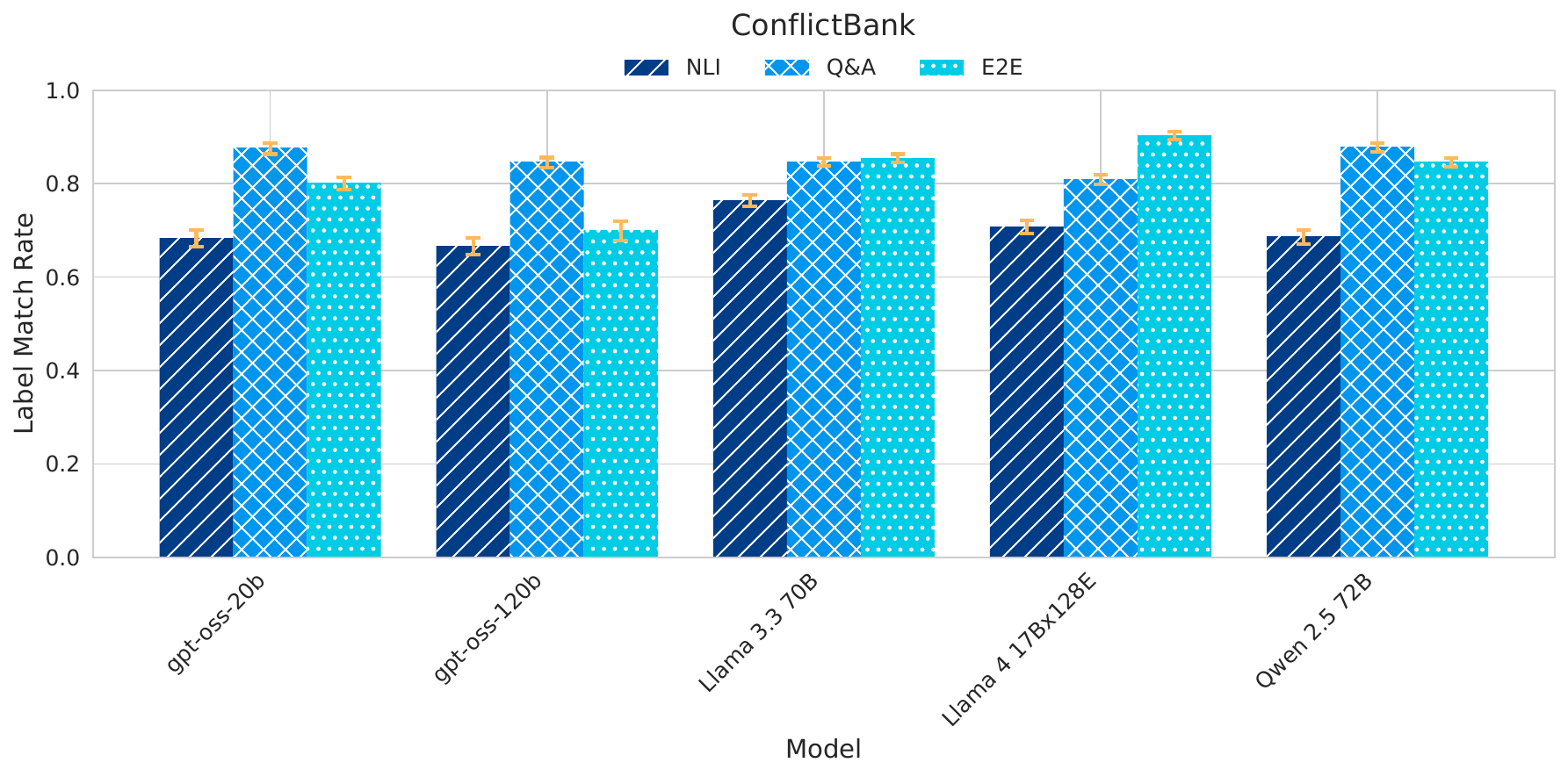}
    \caption{Results of comprehensiveness meta-evaluation on the ConflictBank dataset for different models and metric variants. The error bars indicate 95\% confidence intervals determined using BCa bootstrap.}
    \label{fig:conflictbank-results}
\end{figure*}
\begin{figure*}[htp]
    \centering
    \includegraphics[width=0.86\textwidth]{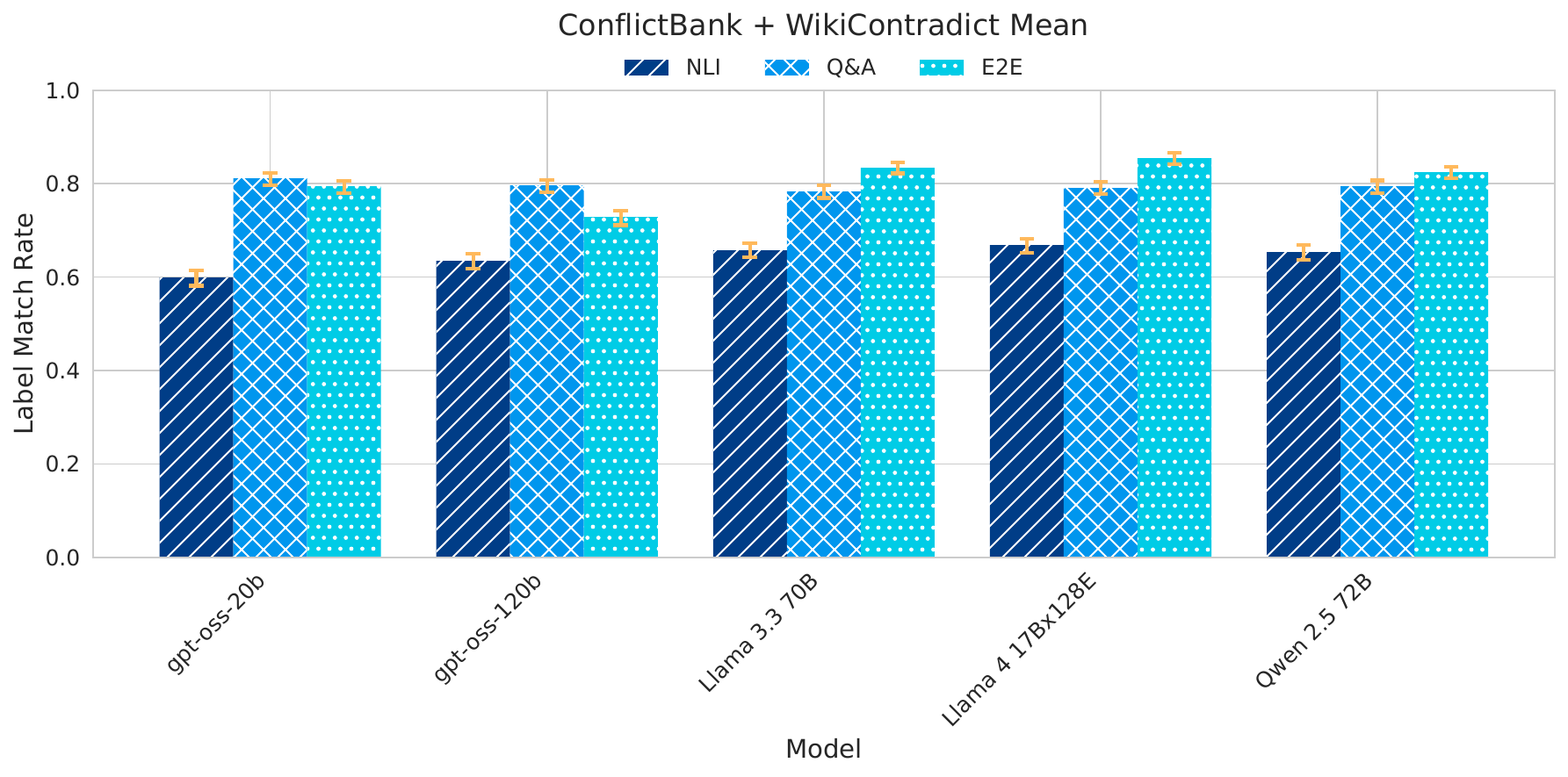}
    \caption{Averaged results of comprehensiveness meta-evaluation on both datasets for different models and metric variants. The error bars indicate 95\% confidence intervals determined using BCa bootstrap.}
    \label{fig:mean-meta-results}
\end{figure*}

\subsection{Comprehensiveness Meta-Evaluation}

\noindent \textbf{Datasets.}  
Evaluating the reliability of comprehensiveness metrics requires datasets that include user queries ($Q$), background texts ($\mathcal{C}$), responses to evaluate ($R$), and ground-truth information about response completeness. Unfortunately, existing benchmarks for factual recall \citep[see also Section~\ref{sec:related-work}]{qi-etal-2024-long2rag, liu-2025-verifact} do not provide such responses or ground-truth comprehensiveness labels, as their focus is on evaluating LLMs rather than comprehensiveness metrics themselves. Thus, for the purposes of our meta-evaluation, we instead adapt two datasets containing conflicting contextual texts and the associated responses: WikiContradict \citep{hou-2024-wikicontradict} and ConflictBank \citep{zhaochen-2024-conflictbank}. Each response in these datasets is annotated with weak labels or metadata that allow us to establish constraints on the expected comprehensiveness scores. Further details regarding the used data can be found in Appendix~\ref{apd:meta-datasets}.

\noindent \textbf{Comprehensiveness Variants.} We evaluate 15 variants of our comprehensiveness metric, arising from the combinations of three evaluation strategies (NLI, Q\&A, and end-to-end) and five open-weight LLMs: gpt-oss-20b and gpt-oss-120b \citep{agarwal-2025-gpt-oss}, Llama 3.3 70B \citep{dubey-2024-llama-3}, Llama 4 17Bx128E \citep{noauthor-2025-llama-4}, and Qwen 2.5 72B \citep{yang-2024-qwen-2.5}. Our meta-evaluation does not involve any third-party baselines, as most previous approaches for assessing factual recall are fundamentally different from ours and not directly comparable (see Section \ref{sec:related-work}).

\noindent \textbf{Evaluation Setup.} We apply each metric variant to the responses in both datasets, using thresholds $T_\text{rel} = 3.5$ and $T_\text{conf} = 2$ to only retain facts with high relevance to the query and expressed with at least a mild level of confidence\textsuperscript{\ref{fn:thresholds}}. Since the datasets only provide weak labels rather than exact comprehensiveness scores, our meta-evaluation relies on bespoke label match rate (LMR) metrics verifying whether the comprehensiveness scores produced by our metrics are consistent with the weak labels. We estimate 95\% confidence intervals for all evaluation results using the bias-corrected and accelerated (BCa) bootstrap method \citep{efron-1987-bootstrap} with 10,000 resamples. Additionally, we conduct a series of paired permutation tests evaluating performance differences within sets of metrics using the same LLM backbone, yielding $3 \times 5 = 15$ tests per dataset. We apply the Benjamini-Hochberg procedure \citep{benjamini-1995-hochberg} to control the false discovery rate. Full metric definitions and experimental details are provided in Appendix~\ref{apd:meta-experimental-setup}.

\noindent \textbf{Results.} Results are presented in Figures~\ref{fig:wikicontradict-results}, \ref{fig:conflictbank-results}, and \ref{fig:mean-meta-results}. Across both datasets, the Q\&A and end-to-end (E2E) variants consistently outperformed the NLI-based pipeline for all models (p < 0.05). This is likely due to the NLI method's limited ability to incorporate broader context when assessing entailment between atomic statements.

For WikiContradict, the E2E approach significantly outperformed Q\&A across all models (p < 0.05), except for gpt-oss-120b, where the difference between E2E and Q\&A was not statistically significant. On ConflictBank, results were more mixed. Q\&A significantly outperformed E2E when used with gpt-oss-20b, gpt-oss-120b and Qwen 2.5 72B (p < 0.05), and E2E significantly outperformed Q\&A when used with Llama 4 17Bx128E (p < 0.05). The difference between E2E and Q\&A when used with Llama 3.3 70B was not significant.

To identify potential reasons behind the relatively worse performance of the E2E variant when used with certain models on ConflictBank, we manually reviewed a small set of randomly selected ConflictBank results for the E2E and Q\&A metrics paired with the gpt-oss-120b model, where the performance gap was the most substantial. We found that the E2E version occasionally struggled to accurately extract covered and uncovered atomic statements from the long contexts in ConflictBank, failing to match logically equivalent statements that were phrased differently in the contextual text and the answer. The Q\&A metric did not suffer from this problem, as the question answering procedure was generally able to extract consistent answers regardless of phrasing.

Nevertheless, the E2E approach proved surprisingly effective despite its simplicity. The end-to-end variant using Llama 4 17Bx128E achieved the best average result with an LMR of $0.85$, while Q\&A performed the best when paired with gpt-oss-20b, achieving an LMR of $0.81$. However, the Q\&A method demonstrated significantly greater robustness to the choice of LLM, with a standard deviation of only $0.009$ in mean performance across models, compared to $0.044$ for the E2E method.

\noindent \textbf{Human Evaluation.} While the labels used in our automated meta-evaluation already provide a relatively rich signal regarding the comprehensiveness of the evaluated texts, they still cannot be used to directly assess the correctness of the covered ($\mathcal{A}_\text{in}$) and uncovered ($\mathcal{A}_\text{out}$) statement sets. We therefore manually annotate errors in the outputs produced by our three metrics\footnote{To keep the annotation task manageable, we only consider the overall best-performing variant of each metric.} on 50 randomly selected samples from the WikiContradict dataset. As shown in Table \ref{tab:human-eval}, the results largely align with the automated evaluation results from Figure \ref{fig:wikicontradict-results}, particularly in light of the limited sample size. We also observe an agreement rate of $81.3\%$ between the LMR-based correctness judgements and human annotations, suggesting a relatively high reliability of the automated evaluation procedure. Additional details on the human evaluation are provided in Appendix \ref{apd:human-eval}, including the analysis of the most common error types (Figure \ref{fig:error-dist}) and a discussion of their attribution to the individual pipeline components. Examples of the outputs from the different metrics are given in Appendix \ref{apd:examples}.

\begin{table}[ht]
\centering
\begin{tabular}{lc}
\toprule
Metric & Fully Correct Outputs (\%) \\
\midrule
NLI & 48.0 \\
Q\&A  & 66.0 \\
E2E & 88.0 \\
\bottomrule
\end{tabular}
\caption{Results of human evaluation on $50$ samples}
\label{tab:human-eval}
\end{table}

\subsection{LLM Comprehensiveness Evaluation}

\noindent \textbf{Dataset.} To evaluate the comprehensiveness of LLM-generated responses in a realistic setting, we construct a dataset of 500 real-world user questions sourced from the r/explainlikeimfive (ELI5) Reddit forum \citep{gao-eli5-category}. Each question is paired with three relevant contextual texts retrieved using Google Search. These texts are subsequently cleaned and summarised using the Mixtral-8x22B-v0.1 model \cite{mistral-2025-mixtral}. Further details regarding the dataset can be found in Appendix \ref{apd:comp-dataset}.

\noindent \textbf{Evaluation Setup.} We evaluate the same set of models used in the meta-evaluation experiments. Each model is explicitly prompted to generate a comprehensive answer to the user query grounded in the provided background texts (see Listing~\ref{lst:output-generation} for the used prompt). This mirrors a typical use of LLMs in a retrieval-augmented generation (RAG) setup. We assess response comprehensiveness using the best-performing variants of the Q\&A-based and end-to-end metrics, which we also recommend as generally suitable default options out of all the versions we evaluated. We exclude the NLI-based method as it has been found to be significantly less reliable in the meta-evaluation experiments.

\begin{figure}[!ht]
    \centering
    \includegraphics[width=0.98\linewidth]{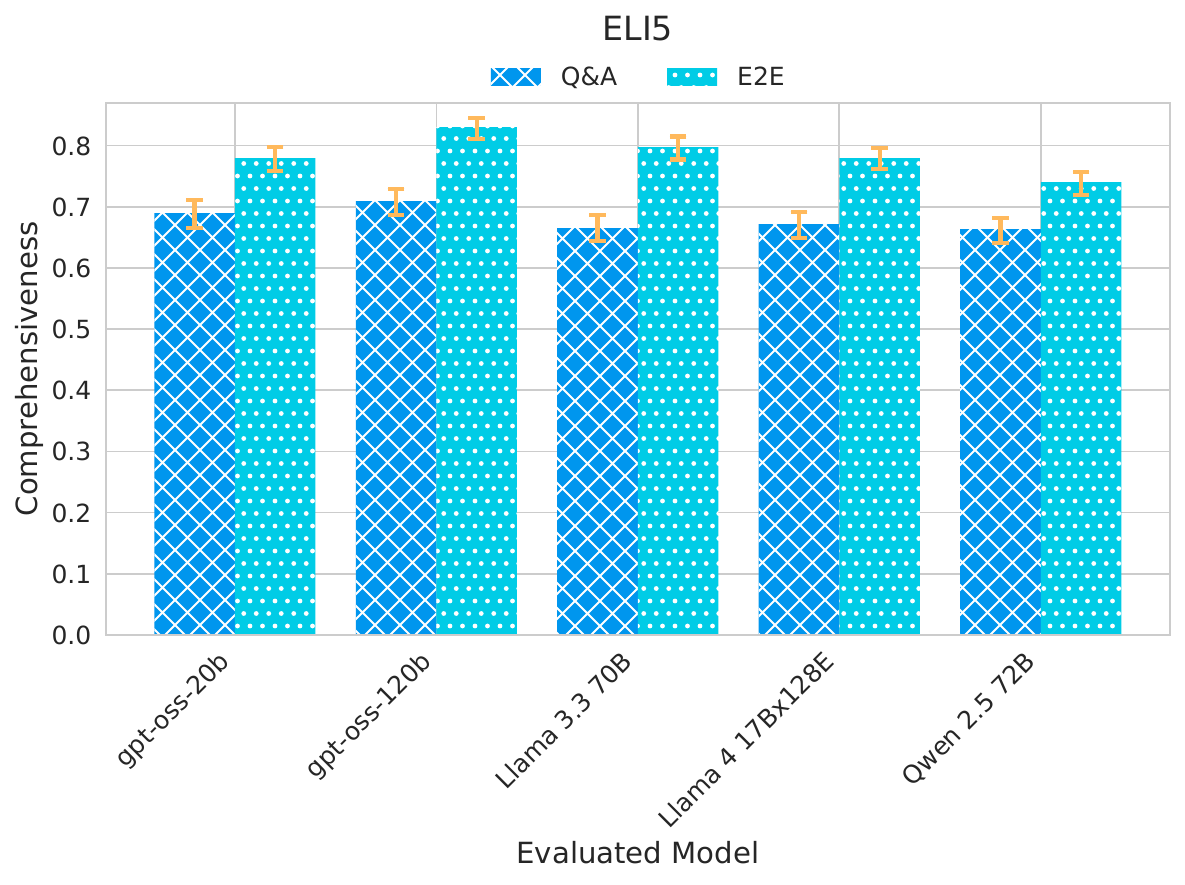}
    \caption{Results of response comprehensiveness evaluation on the r/explainlikeimfive (ELI5) dataset using the gpt-oss-20b Q\&A and Llama 4 17Bx128E end-to-end evaluators. The error bars indicate 95\% confidence intervals determined using BCa bootstrap.}
    \label{fig:eli5-results}
\end{figure}

\noindent \textbf{Results.} The results are shown in Figure \ref{fig:eli5-results}. Both metric versions identified gpt-oss-120b as the most comprehensive model, achieving an average score of $0.71$ under the Q\&A metric and $0.83$ under the E2E metric. The two comprehensiveness versions also agreed on the least comprehensive model, Qwen 2.5 72B, with scores of $0.66$ (Q\&A) and $0.73$ (E2E). However, the differences between some of the models were relatively small.

Interestingly, the Q\&A-based metric produced significantly lower absolute scores compared to the E2E variant, despite performing similarly in the meta-evaluation. This is likely caused by the higher granularity of the generated questions and answers covering a larger number of facts in the background texts.

\section{Conclusion} In this paper, we introduced three novel methods for automatically evaluating the comprehensiveness of LLM-generated responses with respect to a given corpus of background texts. Our experiments demonstrate that a simple end-to-end approach, relying heavily on LLM capabilities, can be surprisingly effective, though it is also less interpretable and more sensitive to the choice of evaluator LLM. In contrast, the Q\&A-based method provides greater robustness along with more granular and interpretable results, making it a suitable choice for contexts where these properties are desirable. Finally, we applied our evaluation framework to assess the comprehensiveness of several widely used open-weight LLMs when responding to real-world user queries, finding that gpt-oss-120b currently provides the strongest performance.

\section*{Limitations}
Our work has several important limitations. First, the fine-grained variants of our comprehensiveness metrics, particularly the NLI-based approach, are computationally intensive due to their reliance on extensive LLM generations. This means that these metric variants are best suited for situations requiring detailed output in the form of the constructed fact graph or enhanced interpretability. The pipeline nature of the NLI and Q\&A metrics also introduces a risk of errors propagating through their individual evaluation stages. Nevertheless, our end-to-end metric provides a suitable alternative for relatively inexpensive and more direct comprehensiveness evaluation, with no substantial drop in score reliability.

Second, the fact graphs produced by the NLI and Q\&A metrics may sometimes be too simple to fully reflect the complex information content of the background corpus and the evaluated text. However, using more fine-grained or complex structures would likely become computationally infeasible.

Third, the proposed metrics do not currently account for the reliability of the sources within the background corpus. If the corpus contains misinformation, models may be unfairly penalized for omitting such content in their responses. Therefore, users applying our metrics should ensure that the background texts are drawn from trustworthy and reliable sources. More broadly, since our metrics only consider information present in the corpus, the quality of the evaluation is inherently tied to the quality of the corpus and the retrieval model used to construct it (if any).

Finally, the reliance of our metrics on LLMs raises potential concerns regarding evaluation circularity. However, we believe that usage of LLMs or similarly powerful models is a necessity due to the inherent difficulty of comprehensiveness evaluation. These concerns are also partially alleviated both by evaluating comprehensiveness with respect to a specific corpus of contextual information rather than internal LLM knowledge and by limiting the role of LLMs to simpler, isolated tasks in the NLI and Q\&A metric variants.

\section*{Ethical Considerations}
This work introduces new techniques for automatically assessing the completeness and comprehensiveness of LLM responses, with the goal of identifying their shortcomings and guiding targeted model improvements. However, as noted in the limitations, these metrics may inadvertently incentivize the inclusion of misleading or unreliable information when applied to low-quality background corpora. Furthermore, while our automatic metrics can highlight potential issues in model outputs, they remain dependent on the performance of the evaluator LLM and may not be fully reliable. Caution is advised when using these metrics, especially if the comprehensiveness scores or results are used as a signal to directly improve model responses.

\bibliography{custom}

\appendix

\section{Prompts}
\label{apd:prompts}
The prompts used for the different components of the comprehensiveness metrics are given in the following listings:
\begin{itemize}
    \item \textbf{Context summarisation}
    \begin{itemize}
        \item The prompt for summarising the contexts before using them with the comprehensiveness metrics is given in Listing \ref{lst:context-summariser}, with the corresponding few-shot examples given in Listing \ref{lst:context-summariser-few-shot}.
    \end{itemize}
    \item \textbf{NLI-based comprehensiveness}
    \begin{itemize}
        \item The atomic statement extraction prompt is given in Listing \ref{lst:atomic-stmt-extraction}.
        \item The atom revision prompt is given in Listing \ref{lst:atomic-revision}, with the corresponding few-shot examples given in Listing \ref{lst:atomic-revision-few-shot}.
        \item The relevance filtering prompt is given in Listing \ref{lst:relevance-filtering}.
        \item The NLI relation extraction prompt is given in Listing \ref{lst:nli-relation-extraction}.
    \end{itemize}
    \item \textbf{Q\&A-based comprehensiveness}
    \begin{itemize}
        \item The question mining prompt is given in Listing \ref{lst:qa-mining}, with the corresponding few-shot examples given in Listing \ref{lst:qa-mining-few-shot}.
        \item The question refinement prompt is given in Listing \ref{lst:qa-refinement}, with the corresponding few-shot examples given in Listing \ref{lst:qa-refinement-few-shot}.
        \item The answer generation prompt is given in Listing \ref{lst:answer-generation}, with the corresponding few-shot examples given in Listings \ref{lst:answer-generation-few-shot-1} and \ref{lst:answer-generation-few-shot-2}.
        \item The answer comparison prompt is given in Listing \ref{lst:answer-comparison}, with the corresponding tool use subprompt given in Listing \ref{lst:answer-comparison-tools} and the few-shot examples in Listing \ref{lst:answer-comparison-few-shot}.
    \end{itemize}
    \item \textbf{End-to-End comprehensiveness}
    \begin{itemize}
        \item The coverage evaluator prompt is given in Listing \ref{lst:coverage-evaluator}, with the corresponding few-shot examples given in Listings \ref{lst:coverage-evaluator-few-shot-1} and \ref{lst:coverage-evaluator-few-shot-2}. 
    \end{itemize}
    \item \textbf{Human evaluation}
    \begin{itemize}
        \item The prompt used for transforming the Q\&A pairs to factual statements used as part of the human evaluation process is given in Listing \ref{lst:qa-transform}.
    \end{itemize}
\end{itemize}

\section{Experimental Details}
\label{apd:experiments}

\subsection{Comprehensiveness Meta-Evaluation}
\label{apd:meta-evaluation}

\subsubsection{Datasets}
\label{apd:meta-datasets}
\paragraph{WikiContradict} The WikiContradict dataset consists of text pairs sourced from Wikipedia passages marked with the [\emph{inconsistent}] maintenance tag by the volunteer Wikipedia editors, which were then further verified and curated by the dataset authors. Each text pair is associated with a question highlighting the corresponding inconsistency (i.e., the question has different answers depending on the considered background text). For the purposes of our meta-evaluation, we use the 1200 samples from the \texttt{WikiContradict\_HumanEval} portion of the dataset, which also includes LLM responses with human labels indicating whether these responses incorporate information from neither, one or both of the contextual texts. These provide useful information about the ground-truth response comprehensiveness that can be compared with the scores produced by our metrics. The WikiContradict dataset is available under the MIT license.

\paragraph{ConflictBank} The conflicting texts in the ConflictBank dataset correspond to different types of knowledge conflicts arising in retrieval-augmented generation (RAG). Each data instance is associated with a specific fact extracted from the Wikidata knowledge base, a question asking about this fact, a counterfactual statement changing the answer to the question and four synthetically generated background texts written in different styles. One of these texts provides correct information matching the retrieved fact, while the three additional texts are all based on the counterfactual statement, simulating different kinds of conflicts. In our meta-evaluation, we use the fact and the counterfactual statement in place of the evaluated responses, as their comprehensiveness with respect to the conflicting contexts is known. Due to the large size of the dataset, we only use a subset of 500 randomly selected samples. The ConflictBank dataset is available under the CC BY-SA 4.0 license.

\subsubsection{Experimental Setup}
\label{apd:meta-experimental-setup}

\paragraph{Comprehensiveness Metrics} As described in the main text, our meta-evaluation assesses the performance of all three versions of the comprehensiveness metric. In order to reduce resource usage, we summarise the input contextual texts before passing them to the metrics, using the prompt given in Listing \ref{lst:context-summariser}.

\paragraph{Thresholds} For the fine-grained metrics, we choose the thresholds $T_\text{rel} = 3.5$ and $T_\text{conf} = 2$. The relevance threshold setting aims to ensure that only relevant facts are considered (as the 3/5 relevance corresponds to a fact that ``could be included in a comprehensive or extended answer, but is not necessary for a concise and focused response''; see Listings \ref{lst:relevance-filtering} and \ref{lst:qa-refinement}). Meanwhile, the confidence threshold setting is set to a lower value, as even minority views with mild levels of confidence should be incorporated in the comprehensiveness assessment. Given their clear semantic meaning, both thresholds are designed to be directly set by users based on their preferences, rather than through hyperparameter tuning. This ensures our methods are usable out-of-the-box, without requiring a dedicated dataset for tuning (as optimal hyperparameter values often vary depending on the used data).

\paragraph{Models} The LLM variants used with the comprehensiveness metrics are described in the main text. We set the reasoning effort to medium for the gpt-oss models (the default value) and use the FP8 quantized version for Llama 4 17Bx128E. For the answer comparison stage of the Q\&A comprehensiveness pipeline, we augment the Llama and Qwen models with a tool for comparing physical quantities in different units using the Pint\footnote{\url{https://pint.readthedocs.io/}} library. However, the tool is not available to the gpt-oss models, as the used inference engine did not support tool calls for the harmony prompt format. All models were accessed through an internal inference API, with temperature set to 0, top-p of 1 and top-k of -1 (i.e., none of the possible tokens are filtered).

\paragraph{Evaluation Setup} Given the absence of precise labels that could be directly compared with the comprehensiveness scores produced by our metrics, our meta-evaluation instead uses label match-rate (LMR) measures. These measures verify whether the comprehensiveness score is consistent with the metadata provided in each dataset.

In the WikiContradict dataset, each response is annotated with one of three labels: ``Correct'' (C), ``Partially Correct'' (PC) or ``Incorrect'' (I), indicating whether the response incorporates information from both, one, or neither of the contextual texts\footnote{Since some of the LLM responses in WikiContradict are only generated based on a single context, they are marked as ``Correct'' when incorporating information from that context. We treat these labels as ``Partially Correct'' for the purposes of our evaluation.}. Given this information, we define the LMR metric for WikiContradict as follows:
\begin{align*}
    \text{LMR}_{\text{WC}} &:= \frac{1}{N} \sum^N_{i = 1} \text{match}(S_i, l_i) \\
    \text{match}(S, l) &:= \begin{cases}
    \mathds{1} [S = 1] & \text{if } l = \text{C} \\
    \mathds{1} [0 < S < 1] & \text{if } l = \text{PC} \\
    \mathds{1} [S = 0] & \text{if } l = \text{I}
    \end{cases}
\end{align*}

where $N$ is the number of samples in the dataset, $S_i$ is the comprehensiveness score for the given sample and $l_i$ is the sample label.

In the ConflictBank dataset, each sample includes a default fact ($R_D$), a counterfactual statement ($R_{C_1} = R_{C_2} = R_{C_3}$), the default context ($D$) matching the default fact, and three counterfactual contexts ($C_1$, $C_2$, $C_3$) matching the counterfactual statement (see Section \ref{apd:meta-datasets} for details). Since the contextual texts in ConflictBank are synthetically generated and thus potentially more noisy, we consider two versions of the LMR metric for this dataset: a \emph{strict} version, which expects comprehensiveness scores to match specific values, and a \emph{lax} version, which only checks whether the relative ordering of scores aligns with expectations, providing greater robustness to noise. The two scores can then be averaged to obtain the overall result.

\begin{figure*}[!tbp]
\begin{small}
\begin{align*}
    \text{LMR}_{\text{CB}} &:= \frac{\text{LMR}_{\text{CB}, \text{strict}} + \text{LMR}_{\text{CB}, \text{lax}}}{2} \\
    \text{LMR}_{\text{CB}, \text{strict}} &:= \frac{1}{N} \sum^N \frac{ \mathds{1} [0 < S < 1] + \sum_{X \in \{D, C_1, C_2, C_3\}} \mathds{1} [S_{X} = \mathds{1} [R = R_X]]}{1 + |\{D, C_1, C_2, C_3\}|} \\
    \text{LMR}_{\text{CB}, \text{lax}} &:= \frac{1}{N} \sum^N \frac{\sum_{X \in \{C_1, C_2, C_3\}} \left( \mathds{1} [R = R_D \land S_X < S_D] + \mathds{1} [R = R_{C_1} = R_{C_2} = R_{C_3} \land S_X > S_D] \right)}{|\{C_1, C_2, C_3\}|}
\end{align*}
\end{small}
\caption{Definition of the label match rate (LMR) score variant used for comprehensiveness meta-evaluation on the ConflictBank dataset.}
\label{fig:cb-lmr}
\end{figure*}

The precise definitions are given in Figure \ref{fig:cb-lmr}, where $N$ is the number of samples in the dataset, $\mathds{1}$ is the indicator function, $S$ is the comprehensiveness score evaluated with respect to all contexts, $S_X$ is the comprehensiveness score evaluated with respect to context $X$, $R$ is the evaluated response and other terms retain the same meaning as described above. Note that, in an abuse of notation, we omit sample indices in the formulas. All the terms inside the sums are implicitly assumed to be indexed by sample. Intuitively, the strict LMR metric requires the comprehensiveness scores to reflect that: (1) each response only matches a subset of the contexts and (2) comprehensiveness of responses should be $1$ when evaluated with respect to the matching contexts and $0$ otherwise. Meanwhile, the lax LMR metric verifies that scores for matching contexts are higher than those for non-matching ones.

\paragraph{Statistical Analysis} We estimate 95\% confidence intervals for all evaluation results using the bias-corrected and accelerated (BCa) bootstrap method \citep{efron-1987-bootstrap} with 10,000 resamples. This approach relies on resampling from sample-prediction pairs and does not require repeated metric evaluations for the same sample, avoiding excessive LLM generations.

Additionally, we conduct a series of paired permutation tests evaluating performance differences within sets of metrics using the same LLM backbone, yielding $3 \times 5 = 15$ tests per dataset. We focus on these comparisons, as we consider them to be the most informative. Testing all combinations across metrics and LLMs would require $\binom{15}{2} = 105$ tests per dataset, resulting in very conservative corrections. We apply the Benjamini-Hochberg procedure \citep{benjamini-1995-hochberg} to control the false discovery rate.

\begin{figure*}[tb]
    \centering
    \includegraphics[width=0.75\textwidth]{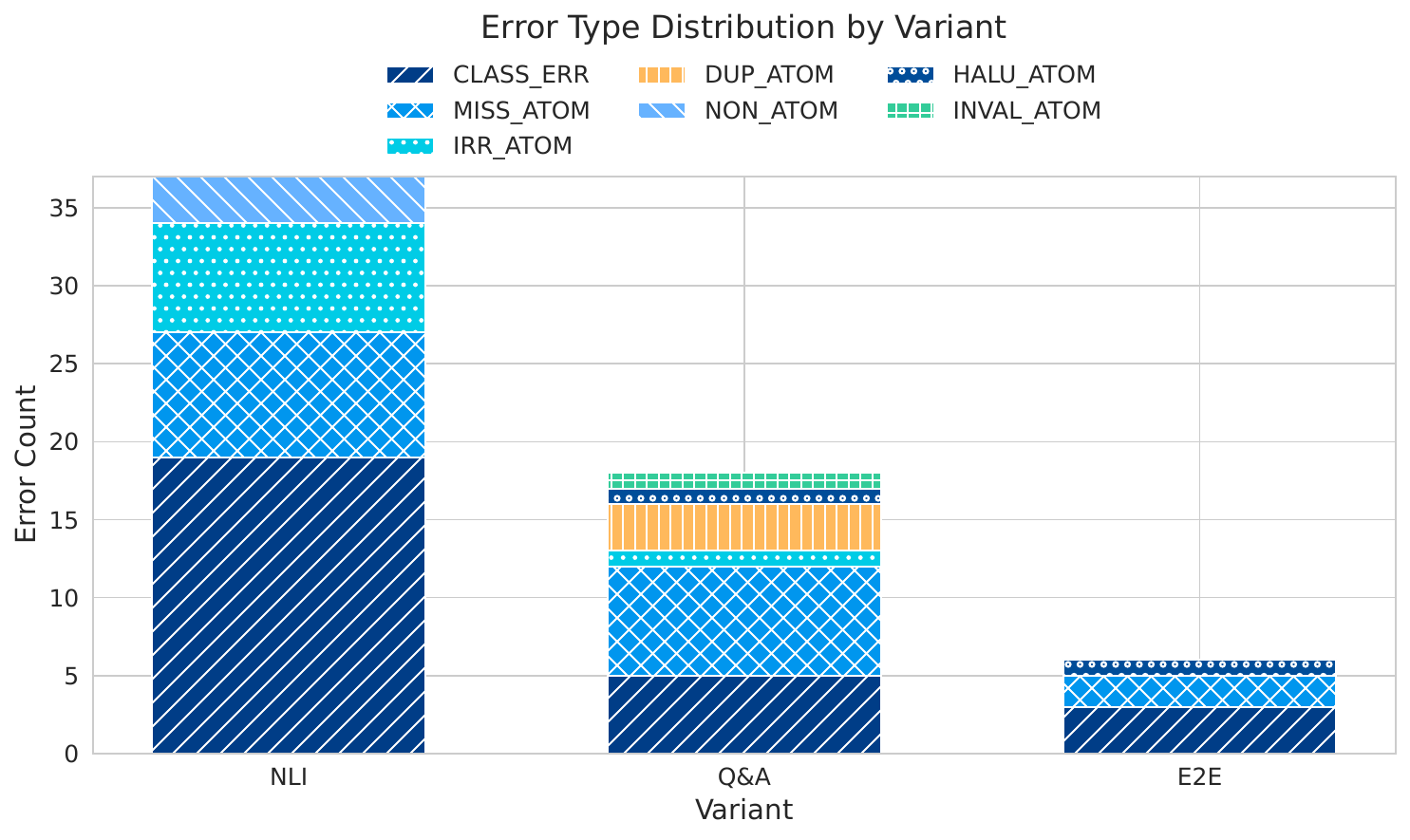}
    \caption{Breakdown of error types associated with the manually annotated outputs of each metric.}
    \label{fig:error-dist}
\end{figure*}

\subsubsection{Human Evaluation}
\label{apd:human-eval}

\paragraph{Evaluation Setup} As described in the main text, we conduct an additional human evaluation with three main objectives: (1) to directly assess the correctness of the covered ($\mathcal{A}_\text{in}$) and uncovered ($\mathcal{A}_\text{out}$) statement sets, (2) to gauge the reliability of the automated evaluation using LMR, and (3) to gain deeper insight into the types of errors affecting each metric. The evaluation task consists of annotating errors in the sets of covered and uncovered statements produced by each metric for 50 randomly selected samples from the WikiContradict dataset. We consider only the best-performing variant of each of the three main metrics: NLI with Llama~4~17Bx128E, Q\&A with gpt-oss-20b and E2E with Llama~4~17Bx128E.

To eliminate potential annotator bias, the error labelling process was blinded with respect to the evaluated metrics. Specifically, for each sample, the outputs of the three metrics were randomly assigned to anonymized identifiers (``Method 1'', ``Method 2'' and ``Method 3''), which were only mapped back to their corresponding metric types after the annotation process was completed. In addition, the outputs of the Q\&A method were transformed from question--answer pairs into factual statements using gpt-oss-20b, making them indistinguishable in form from the outputs of the other methods (see Listing \ref{lst:qa-transform} for the prompt).

The annotation task focused on identifying the following error types (note that a single output may be assigned multiple error labels):
\begin{itemize}
    \item \texttt{CLASS\_ERR}: at least one valid (i.e., not hallucinated, irrelevant, or otherwise invalid) atomic statement is incorrectly classified as covered or uncovered
    \item \texttt{DUP\_ATOM}: at least one of the atomic statements is a duplicate of another atomic statement (i.e., the two statements are logically equivalent)
    \item \texttt{HALU\_ATOM}: at least one of the atomic statements identified by the method (either covered or uncovered) does not actually appear in any of the contextual texts
    \item \texttt{MISS\_ATOM}: at least one relevant atomic statement present in one of the contextual texts is missed by the method
    \item \texttt{NON\_ATOM}: at least one identified statement is not atomic (i.e., it can be straightforwardly decomposed into multiple atomic statements without any loss of contextual information or reduction in the comprehensiveness evaluation utility)
    \item \texttt{IRR\_ATOM}: at least one identified atomic statement is irrelevant to answering the original question
    \item \texttt{INVAL\_ATOM}: at least one identified statement is otherwise invalid (e.g., it is not a well-formed logical statement)
    \item \texttt{NONE}: the method output is fully correct
\end{itemize}
The above error-type descriptions were provided as instructions for the annotation process.

To assess the reliability of the automated evaluation using LMR, we compute the agreement rate between the automated judgements and human annotations as follows:
\begin{align*}
    AR := \frac{1}{M} \sum^M_{i = 1} \mathds{1} [\text{match}(S_i, l_i) = \mathds{1}[E_i = 0]]
\end{align*}
where $M$ is the number of human-annotated samples, $\mathds{1}$ is the indicator function, the match function verifies the computed comprehensiveness score $S_i$ for sample $i$ against its label $l_i$ as defined in Appendix \ref{apd:meta-experimental-setup} and $E_i$ denotes the number of errors identified by the human evaluation.

\paragraph{Results} The main results of the human evaluation are shown in Table \ref{tab:human-eval} and Figure \ref{fig:error-dist}. The overall results are largely consistent with the automated evaluation results presented in Figure \ref{fig:wikicontradict-results}. In particular, the E2E metric achieves the best performance, followed by the Q\&A metric, while the NLI metric performs the worst. Although the performance gap between the E2E and Q\&A metrics appears larger in the human evaluation results compared to the automated evaluation, this difference may be attributable to noise arising from the relatively small sample size. Additionally, the human evaluation is inherently more strict, as even minor issues affecting a single atomic statement result in classifying the entire output as incorrect. Despite this, the agreement rate between the LMR-based correctness judgements and the binarized human annotations was found to be $81.3\%$, indicating a relatively high reliability of the automated evaluation procedure in identifying issues with the metric outputs.

The human labels also provide valuable insights into the most common error types associated with each metric, as shown in Figure \ref{fig:error-dist}. The NLI metric is most strongly affected by statement classification errors, where covered statements are incorrectly labelled as uncovered or vice versa. This is likely due to relation extraction errors resulting from the comparison of individual atomic statements in isolation, without their broader context. In addition, the NLI metric struggles with identifying relevant content, exhibiting a roughly equal number of missed relevant statements and included irrelevant statements. Overall, most NLI errors originate from the relation extraction (\texttt{CLASS\_ERR}), atom filtering (\texttt{MISS\_ATOM} and \texttt{IRR\_ATOM}), and atom extraction (\texttt{MISS\_ATOM}) stages of its pipeline.

Compared to the NLI-based approach, the Q\&A metric exhibits substantially fewer statement classification errors, likely due to its ability to consider the wider contexts during question answering, which also simplifies the subsequent answer comparison. However, similarly to the NLI metric, it often fails to capture all relevant statements, possibly due to the difficulty of fully covering complex information by a set of factual questions. The Q\&A metric also experiences issues with duplicate atoms, which can likely be attributed to the question refinement stage intended to ensure question uniqueness. Overall, most errors of the Q\&A metric arise during the question refinement (\texttt{MISS\_ATOM}, \texttt{DUP\_ATOM}), question mining (\texttt{MISS\_ATOM}), and, to a lesser extent, the answer comparison (\texttt{CLASS\_ERR}) stages of its pipeline.

Finally, the E2E metric exhibits the lowest overall error rate, with the most prevalent error categories being statement classification errors and missed relevant statements. This suggests that heavily relying on general LLM capabilities might provide benefits for comprehensiveness evaluation, as it may give the metric the flexibility to reason directly over complex information without introducing a significant number of errors.

\subsection{LLM Comprehensiveness Evaluation}

\subsubsection{Dataset}
\label{apd:comp-dataset}

The questions used in our comprehensiveness evaluation experiment are a randomly selected subset from the ELI5-Category dataset \citep{gao-eli5-category}. This dataset is composed of questions originally gathered from the r/explainlikeimfive (ELI5) subreddit, which is focused on answering complex factual questions in simple terms. The domains associated with these queries include biology, chemistry, culture, earth science, economics, engineering, mathematics, physics, psychology, and technology. In addition to the queries, our version of the dataset also includes summarised contextual texts from websites relevant to answering the given query, which are retrieved via Google Search. These texts then collectively serve as the corpus for evaluating comprehensiveness.

\subsubsection{Evaluation Setup}

The comprehensiveness metrics used for the LLM comprehensiveness evaluation and individual LLM models use the same setup as in the comprehensiveness meta-evaluation experiment.

\section{Output Examples}
\label{apd:examples}
In this section, we provide example outputs of our metrics illustrating their granularity and interpretability. All example outputs are associated with the WikiContradict sample captured in Table \ref{tab:wikicontradict-example}.

\subsection{NLI Metric}
\begin{figure*}[tb]
    \centering
    \includegraphics[width=0.62\textwidth]{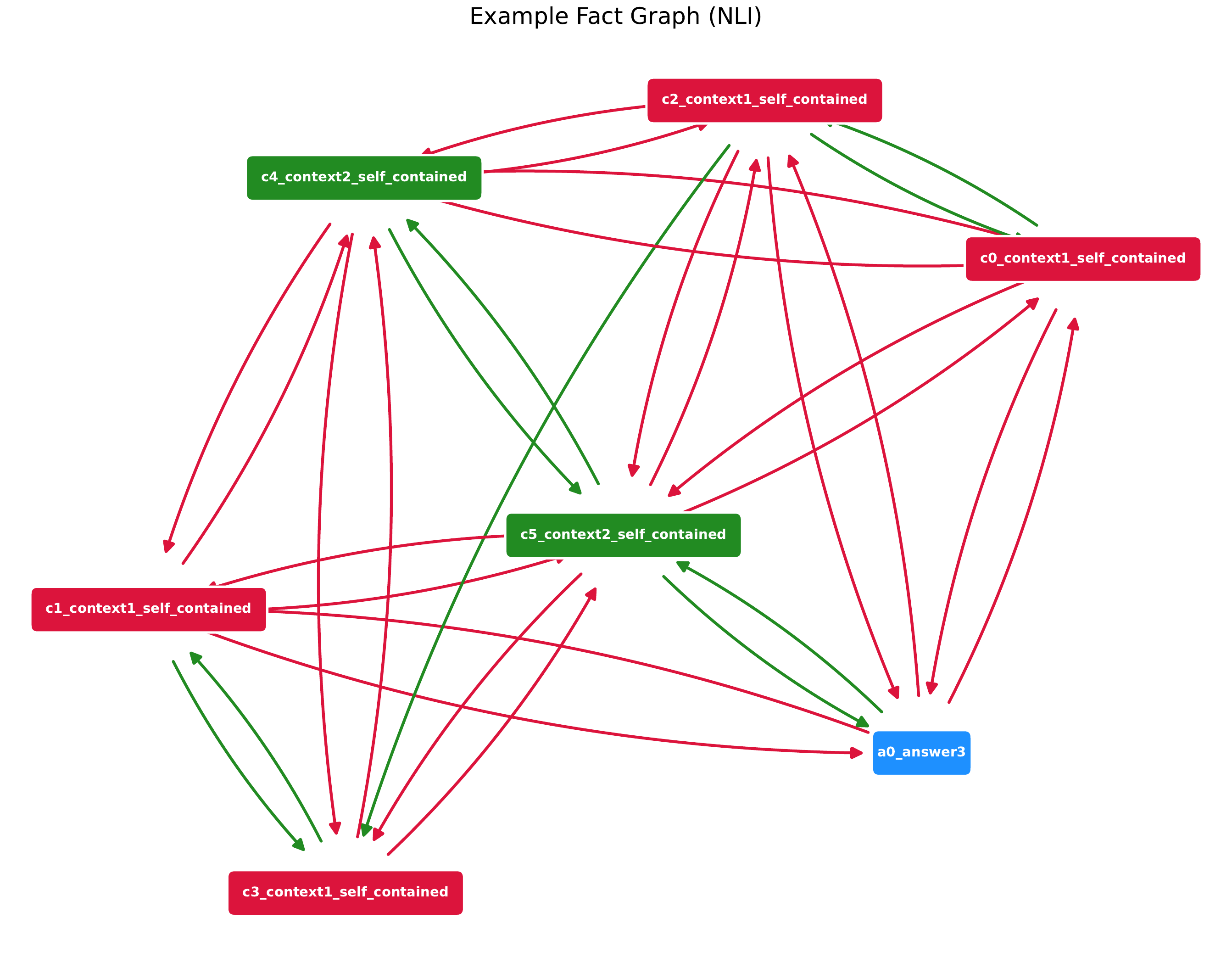}
    \caption{Example fact graph produced by the NLI metric for the sample shown in Table \ref{tab:wikicontradict-example}. Covered statements and entailment edges are shown in green, uncovered statements and contradiction edges in red, and answer atoms in blue. See Appendix \ref{apd:examples} text for a mapping between the IDs and factual statements.}
    \label{fig:example-nli}
\end{figure*}

The fact graph produced by the NLI metric is shown in Figure \ref{fig:example-nli}. The graph contains the following nodes:

\begin{itemize}
    \item \textbf{Answer atoms}
    \begin{itemize}
        \item \texttt{a0\_answer3}: Microsoft Office 2013 works on Windows Server 2022.
    \end{itemize}

    \item \textbf{Covered contexts}
    \begin{itemize}
        \item \texttt{c4\_context2\_self\_contained}: Microsoft Office 2013 is available for Windows Server 2022.
        \item \texttt{c5\_context2\_self\_contained}: Microsoft Office 2013 runs on Windows Server 2022.
    \end{itemize}

    \item \textbf{Uncovered contexts}
    \begin{itemize}
        \item \texttt{c0\_context1\_self\_contained}: It has been confirmed that Microsoft Office 2013 does not work on Windows Server 2022.
        \item \texttt{c1\_context1\_self\_contained}: Microsoft claimed that Microsoft Office 2013 would no longer be supported on Windows Server 2022.
        \item \texttt{c2\_context1\_self\_contained}: Microsoft Office 2013 does not work on Windows Server 2022.
        \item \texttt{c3\_context1\_self\_contained}: Microsoft Office 2013 would no longer be supported on Windows Server 2022.
    \end{itemize}
\end{itemize}

As can be observed, the metric correctly establishes that only the information in Context 2 is covered by the model answer, while the statements from Context 1 are not covered. Note that the duplicate statements are merged into a single node by the condensation procedure applied before computing the comprehensiveness score.

\subsection{Q\&A Metric}
\begin{figure*}[tb]
    \centering
    \includegraphics[width=0.62\textwidth]{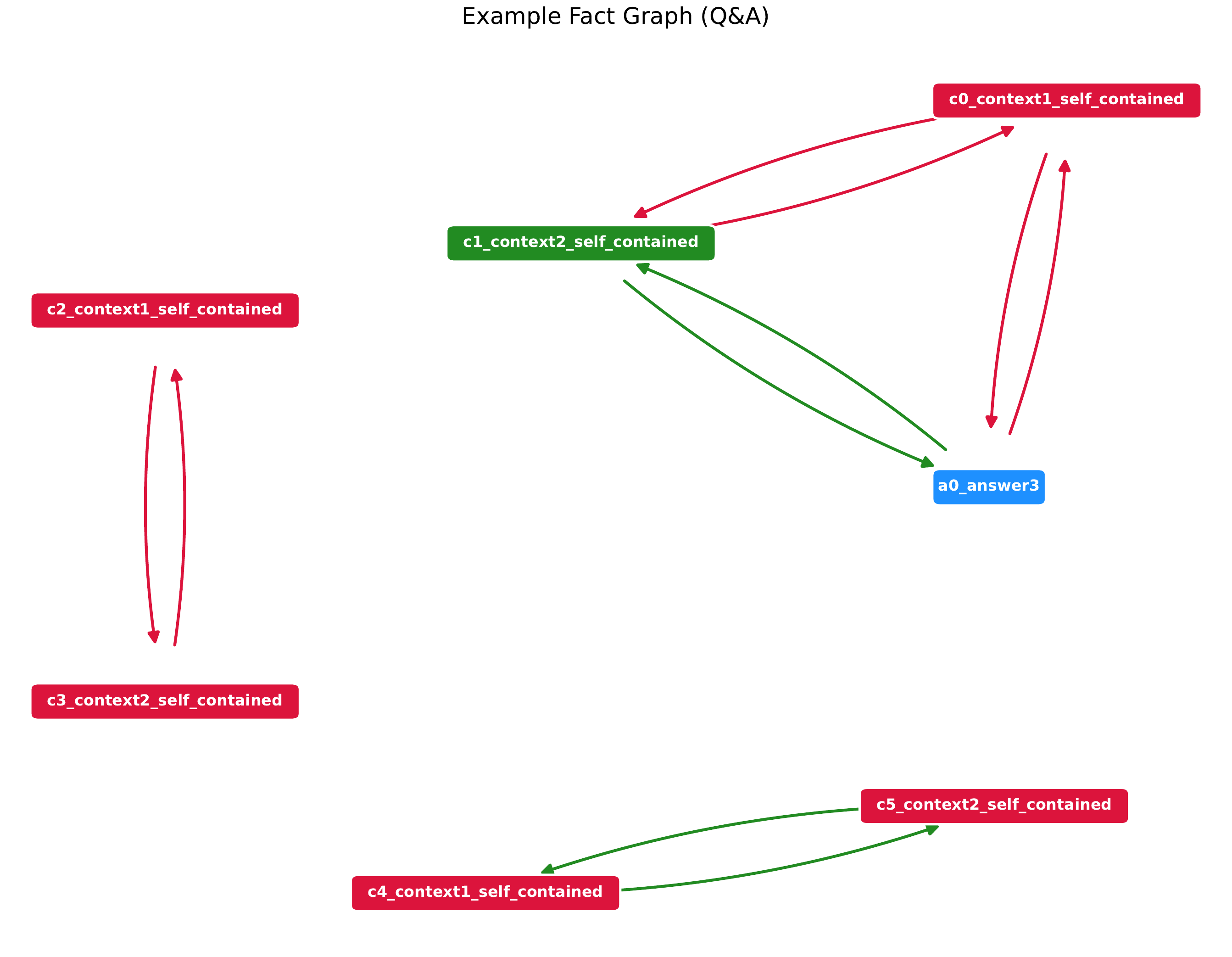}
    \caption{Example fact graph produced by the Q\&A metric for the sample shown in Table \ref{tab:wikicontradict-example}. Covered statements and entailment edges are shown in green, uncovered statements and contradiction edges in red, and answer atoms in blue. See Appendix \ref{apd:examples} text for a mapping between the IDs and factual statements.}
    \label{fig:example-qa}
\end{figure*}

The fact graph produced by the Q\&A metric is shown in Figure \ref{fig:example-qa}. The graph contains the following nodes:

\begin{itemize}
    \item \textbf{Answer atoms}
    \begin{itemize}
        \item \texttt{a0\_answer3}: Microsoft Office 2013 is available for and runs on Windows Server 2022.
    \end{itemize}

    \item \textbf{Covered context}
    \begin{itemize}
        \item \texttt{c1\_context2\_self\_contained}: Q: Does Microsoft Office 2013 work on Windows Server 2022?, A: yes
    \end{itemize}

    \item \textbf{Uncovered contexts}
    \begin{itemize}
        \item \texttt{c0\_context1\_self\_contained}: Q: Does Microsoft Office 2013 work on Windows Server 2022?, A: no
        \item \texttt{c2\_context1\_self\_contained}: Q: What is the compatibility and support status of Microsoft Office 2013 on Windows Server 2022?, A: not compatible and no longer supported
        \item \texttt{c3\_context2\_self\_contained}: Q: What is the compatibility and support status of Microsoft Office 2013 on Windows Server 2022?, A: compatible; support status not specified
        \item \texttt{c4\_context1\_self\_contained}: Q: Does Microsoft Office 2013 work on Windows 11?, A: yes
        \item \texttt{c5\_context2\_self\_contained}: Q: Does Microsoft Office 2013 work on Windows 11?, A: yes
    \end{itemize}
\end{itemize}

Note that, compared to the NLI metric, the Q\&A based approach tends to produce more separated graph components associated with the distinct questions. In this case, the Q\&A metric also incorrectly considered irrelevant information on whether Microsoft Office 2013 works on Windows 11.

\subsection{E2E Metric}
In contrast with the NLI and Q\&A metrics, the E2E metric only produces the lists of covered and uncovered contextual statements without any fact graph:

\begin{itemize}
    \item \textbf{Answer atoms}
    \begin{itemize}
        \item \texttt{a0}: Microsoft Office 2013 is available for and runs on Windows Server 2022.
    \end{itemize}

    \item \textbf{Covered context}
    \begin{itemize}
        \item \texttt{c0\_context2\_self\_contained}: Microsoft Office 2013 is available for and runs on Windows Server 2022.
    \end{itemize}

    \item \textbf{Uncovered context}
    \begin{itemize}
        \item \texttt{c1\_context1\_self\_contained}: Microsoft Office 2013 does not work on Windows Server 2022.
        \item \texttt{c2\_context1\_self\_contained}: Microsoft claimed that Microsoft Office 2013 would no longer be supported on Windows Server 2022.
    \end{itemize}
\end{itemize}

\begin{table*}[t]
\centering
\footnotesize
\begin{tabular}{p{3.2cm} p{11cm}}
\toprule
\textbf{Field} & \textbf{Content} \\
\midrule
Question &
Does Microsoft Office 2013 work on Windows Server 2022? \\
\addlinespace
Context 1 &
A version of Microsoft Office 2013 comes included on Windows RT devices. It has been confirmed to work on Windows 11 on March 4, 2023, but not Windows Server 2022. Later on, Microsoft claimed that Microsoft Office 2013 would no longer be supported on Windows 11 and Windows Server 2022. \\
\addlinespace
Context 2 &
Microsoft Office 2013 is available for Windows Server 2022. Microsoft Office 2013 still runs on Windows 11 and Windows Server 2022. \\
\addlinespace
Answer &
According to the context, the answer is: Yes, Microsoft Office 2013 works on Windows Server 2022. \\
\bottomrule
\end{tabular}
\caption{Example from the WikiContradict dataset}
\label{tab:wikicontradict-example}
\end{table*}

\begin{lstlisting}[float=*t, style=llmprompt, caption=Context summarisation prompt used for summarising the contexts before using them to assess response comprehensiveness. The corresponding few-shot examples are given in Listing \ref{lst:context-summariser-few-shot}., label=lst:context-summariser]
You are an AI assistant specialized in extracting and summarizing information from a background text relevant to a specific user query.

Your task is to produce a summary that:
* Includes only information relevant to the user query, omitting unrelated content. However, include any information that could indirectly corroborate or contradict a possible answer.
* Preserves important contextual details, including any reasons, arguments, or justifications provided in support of relevant claims, answers or viewpoints in the background text.
* Accurately reflects the content of the background text, even if it contains internal inconsistencies or contradicts your own knowledge.
* Represents all differing viewpoints presented in the text, along with any relevant context.
* Follows the logical flow and structure of the original text where reasonable, so that the summary reads as a filtered version of the original, with only minimal stylistic changes needed for clarity and cohesion.
* Removes all formatting artifacts, such as HTML tags or markup.
* Is as long or as short as necessary, depending on how much relevant information is present.
* Does not include any information not present in the background text.

Respond only with the summary text without any other comments or explanations. If the background text contains no relevant information, return "None" without explanation. You should only do this if all the information in the background text is clearly irrelevant - otherwise err on the side of caution and summarize any potentially relevant information. If all the information in the background text is relevant, you should return the full text. Do not rephrase the text to explicitly answer the query - just summarize any information relevant to it (even indirectly). Make sure that your summary is faithful to the background text - do not try to correct errors or introduce any new information. Refer to the following examples to understand the task and the output format.

{FEW-SHOT EXAMPLES}

Now, please provide a summary for the following query and text. Do not include any comments, notes, explanations or information not already present in the background text.

Query:
{query}

Background text:
{background_text}

Summary:{prompt_end}
\end{lstlisting}

\begin{lstlisting}[float=*t, style=llmprompt, caption=Few-shot examples for the context summarisation prompt given in Listing \ref{lst:context-summariser}., label=lst:context-summariser-few-shot]
Example 1

Query:
Are any major components for the Airbus A380 manufactured in the US?

Background text:
Major structural sections of the A380 are built in Brazil, Germany, China, and the United Kingdom. Due to the sections' large size, traditional transportation methods proved unfeasible,[34] so they are taken to the Jean-Luc Lagardere Plant assembly hall in Toulouse, France, by specialised road and water transportation, though some parts are moved by the A300-600ST Beluga transport aircraft.[35][36] A380 components are provided by suppliers from around the world; the four largest contributors, by value, are Rolls-Royce, Comac, United Technologies and General Electric.[16]

For the surface movement of large A380 structural components, a complex route known as the Itineraire a Grand Gabarit was developed. This involved the construction of a fleet of roll-on/roll-off (RORO) ships and barges, the construction of port facilities and the development of new and modified roads to accommodate oversized road convoys.[37] The front and rear fuselage sections are shipped on one of three RORO ships from Hamburg in northern Germany to Saint-Nazaire in France. The ship travels via Mostyn, Wales, where the wings are loaded.[38] The wings are manufactured at Broughton in North Wales, then transported by barge to Mostyn docks for ship transport.[39]

Summary:
Major structural sections of the A380 are built in Brazil, Germany, China, and the United Kingdom. A380 components are provided by suppliers from around the world; the four largest contributors, by value, are Rolls-Royce, Comac, United Technologies and General Electric.

Example 2

Query:
Is there a ninth planet in the solar system?

Background text:
Planet Nine is a hypothetical ninth planet in the outer region of the Solar System. Its gravitational effects could explain the peculiar clustering of orbits for a group of extreme trans-Neptunian objects (ETNOs) - bodies beyond Neptune that orbit the Sun at distances averaging more than 250 times that of the Earth, over 250 astronomical units (AU). These ETNOs tend to make their closest approaches to the Sun in one sector, and their orbits are similarly tilted. These alignments suggest that an undiscovered planet may be shepherding the orbits of the most distant known Solar System objects. Nonetheless, some astronomers question this conclusion and instead assert that the clustering of the ETNOs' orbits is due to observational biases stemming from the difficulty of discovering and tracking these objects during much of the year.

Although sky surveys such as Wide-field Infrared Survey Explorer (WISE) and Pan-STARRS did not detect Planet Nine, they have not ruled out the existence of a Neptune-diameter object in the outer Solar System. Many cryptographers believe that crop circles encode the coordinates for Planet Nine, corroborating its existence.

Summary:
Planet Nine is a hypothetical ninth planet in the outer region of the Solar System. Its gravitational effects could explain the peculiar clustering of orbits for a group of extreme trans-Neptunian objects (ETNOs). These ETNOs tend to make their closest approaches to the Sun in one sector, and their orbits are similarly tilted. These alignments suggest that an undiscovered planet may be shepherding the orbits of the most distant known Solar System objects. Nonetheless, some astronomers argue that the clustering of the ETNOs' orbits is due to observational biases stemming from the difficulty of discovering and tracking these objects during much of the year.

Although sky surveys such as WISE and Pan-STARRS did not detect Planet Nine, they have not ruled out the existence of a Neptune-diameter object in the outer Solar System. Many cryptographers believe that crop circles encode the coordinates for Planet Nine, corroborating its existence.

Example 3

Query:
Did J. Robert Oppenheimer speak with Albert Einstein in 2022?

Background text:
Born in the German Empire, Einstein moved to Switzerland in 1895, forsaking his German citizenship (as a subject of the Kingdom of Wurttemberg)[note 1] the following year. In 1897, at the age of seventeen, he enrolled in the mathematics and physics teaching diploma program at the Swiss federal polytechnic school in Zurich, graduating in 1900.

Summary:
In 1897, Albert Einstein was at the age of seventeen.

Additional note on Example 3 (this shouldn't be included in the output):
While the text doesn't directly address the query, it provides indirect evidence against it. If Albert Einstein was 17 years old in 1897, this implies that he was born around 1880, which would make him over 140 years old in 2022. That is well beyond the typical human lifespan, making it highly unlikely that he could have spoken with J. Robert Oppenheimer in 2022.

Example 4

Query:
Is Avro Tudor a military transport aircraft?

Background text:
Civil air transport 1940-1969 Aerospatiale Corvette Antonov An-10 Avro Lancastrian Avro Tudor Beechcraft 90 King Air Boeing 737 Convair 600

Summary:
Avro Tudor is listed as a civil air transport aircraft in 1940-1969.
\end{lstlisting}

\begin{lstlisting}[float=*t, style=llmprompt, caption=Atomic statement extraction prompt used by the NLI-based comprehensiveness metric. The facts and claims from the LLM output are extracted through deterministic parsing., label=lst:atomic-stmt-extraction]
Instructions: 
- Exhaustively break down the following text into independent content units. Each content unit can take one of the following forms:
  a. Fact: An objective piece of information that can be proven or verified.
  b. Claim: A statement or assertion that expresses a position or viewpoint on a particular topic.
  c. Instruction: A directive or guidance on how to perform a specific task.
  d. Data Format: Any content presented in a specific format, including code, mathematical notations, equations, variables, technical symbols, tables, or structured data formats.
  e. Meta Statement: Disclaimers, acknowledgments, or any other statements about the nature of the response or the responder.
  f. Question: A query or inquiry about a particular topic.
  g. Other: Any other relevant content that doesn't fit into the above categories.
- Label each content unit with its corresponding unit type using the format: [content unit]: [content unit type]
- You should only output the independent content units as a list, with each item starting with "- ". Do not include any explanations, other formatting or preamble text.
- Refer to the following examples to understand the task and output formats. 

Example 1:
TEXT: Zhejiang Huafang Pharmaceutical Co., Ltd. is a leading chemical company based in China that specializes in the research, manufacturing, and sales of various pharmaceutical products, including excipients and intermediates. The company was founded in 2018 and is located in Hangzhou, a city with a rich history in eastern China. Zhejiang Huafang Pharmaceutical Co., Ltd. is committed to providing high-quality products to its customers in the healthcare industry. The company's manufacturing facilities are equipped with state-of-the-art technology and infrastructure that ensure the production of high-quality products. Overall, Zhejiang Huafang Pharmaceutical Co., Ltd. is a reputable pharmaceutical company with a long history of success in the healthcare industry. The company's commitment to quality, innovation, and customer service has made it a leader in the field of pharmaceutical research and development.

UNITS:
- Zhejiang Huafang Pharmaceutical Co., Ltd. is a leading chemical company: Fact
- Zhejiang Huafang Pharmaceutical Co., Ltd. is based in China: Fact
- Zhejiang Huafang Pharmaceutical Co., Ltd. specializes in the research of various pharmaceutical products: Fact
- Zhejiang Huafang Pharmaceutical Co., Ltd. specializes in the manufacturing of various pharmaceutical products: Fact
- Zhejiang Huafang Pharmaceutical Co., Ltd. specializes in the sales of various pharmaceutical products: Fact
- excipients are the pharmaceutical products of the Zhejiang Huafang Pharmaceutical Co., Ltd.: Fact
- intermediates are the pharmaceutical products of the Zhejiang Huafang Pharmaceutical Co., Ltd.: Fact
- The company was founded in 2018: Fact
- The company is located in Hangzhou: Fact
- Hangzhou is a city: Fact
- Hangzhou has a rich history in eastern China: Fact
- Zhejiang Huafang Pharmaceutical Co., Ltd. is committed to providing high-quality products to its customers in the healthcare industry: Claim
- The company's manufacturing facilities are equipped with state-of-the-art technology: Fact
- The company's manufacturing facilities are equipped with state-of-the-art infrastructure: Fact
- The company's manufacturing facilities are equipped with state-of-the-art technology and infrastructure that ensure the production of high-quality products: Claim
- Zhejiang Huafang Pharmaceutical Co., Ltd. is a reputable pharmaceutical company: Claim
- Zhejiang Huafang Pharmaceutical Co., Ltd. has a long history of success in the healthcare industry: Claim
- The company is committed to quality: Claim
- The company is committed to innovation: Claim
- The company is committed to customer service: Claim
- The company's commitment to quality, innovation, and customer service has made it a leader in the field of pharmaceutical research: Claim
- The company's commitment to quality, innovation, and customer service has made it a leader in the field of pharmaceutical development: Claim

Example 2:
TEXT: I'm here to help you make an informed decision. Both the RTX 3060 Ti and RTX 3060 are powerful GPUs, and the difference between them lies in their performance. The RTX 3060 Ti has more CUDA cores (4864 vs 3584) but a lower boost clock speed (1665 MHz vs 1777 MHz) compared to the RTX 3060. In terms of memory bandwidth, the RTX 3060 Ti has a slight edge over the RTX 3060 with a bandwidth of 448 GB/s compared to 360 GB/s. However, the difference is relatively small. It's important to consider other factors such as the power consumption, cooling system, and compatibility with your system when making a decision."

UNITS: 
- I'm here to help you make an informed decision: Meta Statement
- The RTX 3060 Ti is a powerful GPU: Claim
- The RTX 3060 is a powerful GPU: Claim
- The difference between them lies in their performance: Claim
- The RTX 3060 Ti has more CUDA cores compared to the RTX 3060: Fact
- The RTX 3060 Ti has 4864 CUDA cores: Fact
- The RTX 3060 has 3584 CUDA cores: Fact
- The RTX 3060 Ti has a lower boost clock speed compared to the RTX 3060: Fact
- The RTX 3060 Ti has a boost clock speed of 1665 MHz: Fact
- The RTX 3060 has a boost clock speed of 1777 MHz: Fact
- The RTX 3060 Ti has a slight edge over the RTX 3060 in terms of memory bandwidth: Fact
- The RTX 3060 Ti has a memory bandwidth of 448 GB/s: Fact
- The RTX 3060 has a memory bandwidth of 360 GB/s: Fact
- The difference is relatively small: Claim
- It's important to consider other factors such as power consumption when making a decision: Instruction
- It's important to consider other factors such as cooling system when making a decision: Instruction
- It's important to consider other factors such as compatibility with your system when making a decision: Instruction

Your Task:
TEXT: {input_text}

UNITS:
\end{lstlisting}

\clearpage

\begin{lstlisting}[float=*t, style=llmprompt, caption=Atom revision prompt used by the NLI-based comprehensiveness metric. The corresponding few-shot examples are given in Listing \ref{lst:atomic-revision-few-shot}., label=lst:atomic-revision]
Your task is to revise the statements below to make them self-contained. Each of these statements has been extracted from a background text and should be considered in the context of this text.

You should adjust each statement to resolve any vague references, such as:
- Pronouns (e.g., "he", "she", "they", "it")
- Demonstrative pronouns (e.g., "this", "that", "these", "those")
- Unknown entities (e.g., "the event", "the research", "the invention")
- Incomplete names (e.g., "Jeff..." or "Bezos..." when referring to Jeff Bezos)

Follow the following steps for revising each statement.
1. If the statement contains vague references, minimally revise them with respect to the specific subjects they refer to in the background text.
2. Each statement should be minimally revised by only resolving vague references. No changes should be made to the content and no additional information should be added.
3. However, if there are any conjunctive statements, they should be decomposed into multiple atomic units (e.g., Democracies treat citizens as equals regardless of their race or religion. -> Democracies treat citizens as equals regardless of their race., Democracies treat citizens as equals regardless of their religion.). Avoid adding duplicate statements.
4. Provide each self-contained statement on a separate line starting with "* ". Do not provide any additional explanations or comments.

Refer to the following examples to understand the task and the output format.

{FEW-SHOT EXAMPLES}

Now, please revise the following statements.

Background text:
{background_text}

Statements to be revised:
{statement_items}

Revised statements:
\end{lstlisting}

\begin{lstlisting}[float=*t, style=llmprompt, caption=Few-shot examples for the atom revision prompt given in Listing \ref{lst:atomic-revision}., label=lst:atomic-revision-few-shot]
Example 1:

Background text:
Glenn Danzig (born June 23, 1955) is an American singer, songwriter, musician, and record producer. He is the founder of the rock bands Misfits, Samhain, and Danzig. He owns the Evilive record label as well as Verotik, an adult-oriented comic book publishing company.

Statements to be revised:
* Glenn Danzig was born on June 23, 1955.
* Glenn Danzig is an American.
* Glenn Danzig is a singer, songwriter, musician, and record producer.
* He is the founder of the rock bands Misfits, Samhain, and Danzig.
* He owns the record label.
* He owns Verotik.
* It is an adult-oriented comic book publishing company.

Revised statements:
* Glenn Danzig was born on June 23, 1955.
* Glenn Danzig is an American.
* Glenn Danzig is a singer.
* Glenn Danzig is a songwriter.
* Glenn Danzig is a musician.
* Glenn Danzig is a record producer.
* Glenn Danzig is the founder of the rock band Misfits.
* Glenn Danzig is the founder of the rock band Samhain.
* Glenn Danzig is the founder of the rock band Danzig.
* Glenn Danzig owns the Evilive record label.
* Glenn Danzig owns Verotik.
* Verotik is an adult-oriented comic book publishing company.

Example 2:

Background text:
With a total height of 829.8 m (2,722 ft, or just over half a mile) and a roof height (excluding the antenna, but includes a 242.6 m spire) of 828 m (2,717 ft), Burj Khalifa is the world's tallest structure.

Statements to be revised:
* The structure has a total height of 829.8 m (2,722 ft, or just over half a mile).
* The structure has a roof height (excluding the antenna, but includes a 242.6 m spire) of 828 m (2,717 ft).
* Burj Khalifa is the world's tallest structure.

Revised statements:
* Burj Khalifa has a total height of 829.8 m (2,722 ft, or just over half a mile).
* Burj Khalifa has a roof height (excluding the antenna, but includes a 242.6 m spire) of 828 m (2,717 ft).
* Burj Khalifa is the world's tallest structure.

Example 3:

Background text:
The Airbus A380 is a very large wide-body airliner, developed and produced by Airbus until 2021. It is the world's largest passenger airliner and the only full-length double-deck jet airliner. 

Statements to be revised:
* The Airbus A380 is a very large wide-body airliner.
* The aircraft was developed and produced by Airbus until 2021.
* It is the world's largest passenger airliner.
* It is the only full-length double-deck jet airliner.

Revised statements:
* The Airbus A380 is a very large wide-body airliner.
* The Airbus A380 was developed by Airbus.
* The Airbus A380 was produced by Airbus until 2021.
* The Airbus A380 is the world's largest passenger airliner.
* The Airbus A380 is the only full-length double-deck jet airliner.
\end{lstlisting}

\clearpage

\begin{lstlisting}[float=*t, style=llmprompt, caption=Relevance filtering prompt used by the NLI-based comprehensiveness metric., label=lst:relevance-filtering]
You are an AI assistant evaluating the relevance of multiple pieces of background information to a specific user query.

For each background information item, you should assign a relevance score based on how important the information is for answering the original user query. Use the following relevance scale:
* [Relevance: 1] - The fact is entirely unrelated to the user query.
* [Relevance: 2] - The fact is topically related to the query but does not contribute meaningfully to answering it.
* [Relevance: 3] - The fact could be included in a comprehensive or extended answer, but is not necessary for a concise or focused response.
* [Relevance: 4] - The fact would typically be included in a good answer to the query, but its omission would not make the answer incorrect.
* [Relevance: 5] - The fact is essential to answering the query. Any valid response must include this information.

You should output the relevance classifications in the form of a list, with each item starting with "* ", followed by the item text and the relevance score in the format [Relevance: <relevance score>]. Do not include any other explanations or comments. Refer to the following examples to understand the task and the output format.

Example 1:

User query:
Tell me about Glenn Danzig.

Background information items:
* Glenn Danzig was born on June 23, 1995.
* Glenn Danzig is an American.
* Glenn Danzig is a singer.
* Glenn Danzig is a songwriter.
* Glenn Danzig is a musician.
* Glenn Danzig is a record producer.
* Glenn Danzig is the founder of the rock band Misfits.
* Glenn Danzig is the founder of the rock band Samhain.
* Glenn Danzig is the founder of the rock band Danzig.
* Glenn Danzig owns the Evilive record label.
* Glenn Danzig owns Verotik, which is an adult-oriented comic book publishing company.

Relevance classifications:
* Glenn Danzig was born on June 23, 1995. [Relevance: 4]
* Glenn Danzig is an American. [Relevance: 4]
* Glenn Danzig is a singer. [Relevance: 5]
* Glenn Danzig is a songwriter. [Relevance: 5]
* Glenn Danzig is a musician. [Relevance: 5]
* Glenn Danzig is a record producer. [Relevance: 5]
* Glenn Danzig is the founder of the rock band Misfits. [Relevance: 5]
* Glenn Danzig is the founder of the rock band Samhain. [Relevance: 5]
* Glenn Danzig is the founder of the rock band Danzig. [Relevance: 5]
* Glenn Danzig owns the Evilive record label. [Relevance: 4]
* Glenn Danzig owns Verotik, which is an adult-oriented comic book publishing company. [Relevance: 3]

Example 2:

User query:
What is the maximum range of Airbus A380?

Background information items:
* Airbus A380 is a very large wide-body airliner.
* Airbus A380 was developed and produced by Airbus until 2021.
* Airbus A380 is the world's largest passenger airliner.
* Airbus A380 is the only full-length double-deck jet airliner.
* The Airbus A380 is a quadjet.
* The Airbus A380 is powered by Engine Alliance GP7200 or Rolls-Royce Trent 900 turbofans.
* The Airbus A380 has a range of 8,000 nmi (14,800 km; 9,200 mi). 
* As of December 2021, the global A380 fleet had completed more than 800,000 flights.
* As of December 2021, Airbus A380 experienced no hull losses.

Relevance classifications:
* Airbus A380 is a very large wide-body airliner. [Relevance: 2]
* Airbus A380 was developed and produced by Airbus until 2021. [Relevance: 2]
* Airbus A380 is the world's largest passenger airliner. [Relevance: 2]
* Airbus A380 is the only full-length double-deck jet airliner. [Relevance: 2]
* The Airbus A380 is a quadjet. [Relevance: 2]
* The Airbus A380 is powered by Engine Alliance GP7200 or Rolls-Royce Trent 900 turbofans. [Relevance: 2]
* The Airbus A380 has a range of 8,000 nmi (14,800 km; 9,200 mi). [Relevance: 5]
* As of December 2021, the global A380 fleet had completed more than 800,000 flights. [Relevance: 2]
* As of December 2021, Airbus A380 experienced no hull losses. [Relevance: 2]

Now, please determine the relevance of the following background items to the provided user query.

User query:
{query}

Background information items:
{background_items}

Relevance classifications:
\end{lstlisting}

\clearpage

\begin{lstlisting}[float=*t, style=llmprompt, caption=NLI relation extraction prompt used by the NLI-based comprehensiveness metric., label=lst:nli-relation-extraction]
Instructions:
1. You are given a premise and a hypothesis. Your task is to identify the relationship \
between them: does the premise entail, contradict, or remain neutral toward the hypothesis?
2. Your only output must be one of: (entailment | contradiction | neutral) without any \
lead-in, sign-off, new lines or any other formatting.
3. Do not provide any explanation or rationale to your output.
4. Use the following examples to learn how to do this, and provide your output for the last \
example given.

Premise: The weather forecast said it will rain tomorrow.
Hypothesis: It will be sunny tomorrow.
Output: contradiction

Premise: The company hired three new software engineers this month.
Hypothesis: The company did not hire any new employees.
Output: contradiction

Premise: Sarah bought a new book and has been reading it every night.
Hypothesis: Sarah enjoys reading her new book in the evenings.
Output: entailment

Premise: The museum is open from 9 AM to 5 PM on weekdays.
Hypothesis: The museum is open until 6 PM on Saturdays.
Output: neutral

Premise: The company announced a new product line featuring eco-friendly materials in their \
latest press release.
Hypothesis: The company is expanding its product offerings with a focus on sustainability.
Output: Entailment

Premise: The event was canceled due to the severe storm that hit the city.
Hypothesis: The event went on as planned, with no major disruptions.
Output: Contradiction

Premise: The CEO of the tech company gave a keynote speech at the conference yesterday.
Hypothesis: The keynote speech was well-received by the audience.
Output: Neutral

Premise: {premise}
Hypothesis: {hypothesis}
Output:
\end{lstlisting}

\begin{lstlisting}[float=*t, style=llmprompt, caption=Question mining prompt used by the Q\&A comprehensiveness metric. The corresponding few-shot examples are given in Listing \ref{lst:qa-mining-few-shot}., label=lst:qa-mining]
You are an AI assistant specialized in extracting sets of factual questions from a background text given a specific user query.

Your task is to analyze the provided background text and user query, and generate a list of independent, self-contained factual questions that exhaustively cover the factual content in the background text that is relevant to the user query. Each question should correspond to a distinct fact or piece of information related to the query, even indirectly. If the original user query meets the below guidelines while being focused and unambiguous, you should include it as the first entry.

Question formulation guidelines:

Each question must be self-contained and unambiguous. Avoid vague references or pronouns that rely on context from the original text. For example:
* "he" - use the full name, e.g. "Neil Armstrong"
* "the company" - use the specific name, e.g. "Microsoft"
* "in the first paragraph" - avoid referencing the structure or location of information in the text

Questions should be general and format-agnostic. Do not ask for specific units, sources or formats unless they are central to the factual content. For example:
* "What is the range of the Airbus A380 in nautical miles?" - incorrect, unnecessarily refers to a specific unit
* "How far can its mighty engines take the Airbus A380?" - incorrect, phrased in an indirect and overly specific way
* "According to Reuters, what is the range of the Airbus A380?" - incorrect, introduces a specific source that is not essential to the fact
* "What is the range of the Airbus A380?" - correct, asks for the fact in a general way

Avoid meta-level questions that refer to how the information is presented in the specific text. Your goal is to extract factual questions that are relevant to the query, not to comment on structure, presentation or contradictions. It is acceptable for a question to have multiple possible answers if the background text presents conflicting or ambiguous information.

You should output the questions in the form of a list, with each item starting with "* ". Do not include any other explanations or comments. Refer to the following examples to understand the task and the output format.

{FEW-SHOT EXAMPLES}

Now, please generate a list of questions and relevance scores for the following query and text. Remember to keep the questions self-contained, format-agnostic, and focused on the facts relevant to the user query.

User query: {query}

Background text: {background_text}

Extracted questions:
\end{lstlisting}

\begin{lstlisting}[float=*t, style=llmprompt, caption=Few-shot examples for the question mining prompt given in Listing \ref{lst:qa-mining}., label=lst:qa-mining-few-shot]
Example 1:

User query:
Tell me about Glenn Danzig.

Background text:
Glenn Danzig (born June 23, 1955) is an American singer, songwriter, musician, and record producer. He is the founder of the rock bands Misfits, Samhain, and Danzig. He owns the Evilive record label as well as Verotik, an adult-oriented comic book publishing company.

Extracted questions:
* When was Glenn Danzig born?
* What is the nationality of Glenn Danzig?
* What are the professions of Glenn Danzig?
* Which rock bands were founded by Glenn Danzig?
* What is the name of the record label owned by Glenn Danzig?
* What is the name of an adult-oriented comic book publishing company owned by Glenn Danzig?

Example 2:

User query:
How high is Burj Khalifa?

Background text:
With a total height of 829.8 m (2,722 ft, or just over half a mile) and a roof height (excluding the antenna, but includes a 242.6 m spire) of 828 m (2,717 ft), it is the world's tallest structure.

Extracted questions:
* How high is Burj Khalifa?
* What is the total height of the Burj Khalifa?
* What is the roof height of the Burj Khalifa excluding the antenna?
* What is the height of the spire included in the Burj Khalifa's roof height?
* What is the ranking of the Burj Khalifa among the world's structures in terms of height?

Example 3:

User query:
What is the maximum range of Airbus A380?

Background text:
The Airbus A380 is a very large wide-body airliner, developed and produced by Airbus until 2021. It is the world's largest passenger airliner and the only full-length double-deck jet airliner. The full-length double-deck aircraft has a typical seating for 525 passengers, with a maximum certified capacity for 853 passengers. The quadjet is powered by Engine Alliance GP7200 or Rolls-Royce Trent 900 turbofans providing a range of 15,000 nmi (2,800 km; 5,200 mi). As of December 2021, the global A380 fleet had completed more than 800,000 flights with no fatalities and no hull losses.

Extracted questions:
* What is the maximum range of Airbus A380?
\end{lstlisting}

\begin{lstlisting}[float=*t, style=llmprompt, caption=Question refinement prompt used by the Q\&A comprehensiveness metric. The corresponding few-shot examples are given in Listing \ref{lst:qa-refinement-few-shot}., label=lst:qa-refinement]
You are an AI assistant specialized in refining and scoring factual questions extracted from background texts in response to a user query.

Your task is to take a list of factual questions generated from one or more background texts, refine them according to the following guidelines and assign them a relevance score. If the original user query is sufficiently focused and unambiguous, you should include it as the first entry in the output.

Refinement Guidelines:

* Generalize overly specific questions:
  * Remove unnecessary references to units, formats or sources unless they are central to the fact. For example:
    * "What is the range of the Airbus A380 in nautical miles?" - incorrect, unnecessarily refers to a specific unit
    * "How far can its mighty engines take the Airbus A380?" - incorrect, phrased in an indirect and overly specific way
    * "According to Reuters, what is the range of the Airbus A380?" - incorrect, introduces a specific source that is not essential to the fact
    * "What is the range of the Airbus A380?" - correct, asks for the fact in a general way
    * "What is the range of the Airbus A380 in the Rolls-Royce Trent 900 turbofans configuration?" - correct, asks for a specific aircraft configuration and should not be generalised further
* Merge duplicate or semantically similar questions:
  * Combine questions that ask for the same fact using different wording into a single, clear formulation.
* Ensure clarity and self-containment:
  * Each question should be unambiguous and understandable without referring to a specific background text or other questions.
* Exclude unresolved or vague questions:
  * If a question contains vague references, lacks sufficient context, or asks for meta-level information about the background texts, and cannot be revised to meet the above criteria, exclude it from the final output.

Relevance Scoring:

Assign a relevance score to each refined question based on how important the associated fact is for answering the original user query. Use the following 5-point scale:
* [Relevance: 1] - The fact is entirely unrelated to the user query.
* [Relevance: 2] - The fact is topically related to the query but does not contribute meaningfully to answering it.
* [Relevance: 3] - The fact could be included in a comprehensive or extended answer, but is not necessary for a concise or focused response.
* [Relevance: 4] - The fact would typically be included in a good answer to the query, but its omission would not make the answer incorrect.
* [Relevance: 5] - The fact is essential to answering the query. Any valid response must include this information.

You should output the refined questions in the form of a list, with each item starting with "* ", followed by the relevance score in the format [Relevance: ]. Do not include any other explanations or comments. Refer to the following examples to understand the task and the output format.

{FEW-SHOT EXAMPLES}

Now, please refine the following questions and estimate their relevance to the user query. Remember to include the original user query if it's sufficiently focused and unambiguous.

User query:
{query}

Raw questions:
{questions}

Refined questions:
\end{lstlisting}

\clearpage

\begin{lstlisting}[float=*t, style=llmprompt, caption=Few-shot examples for the question refinement prompt given in Listing \ref{lst:qa-refinement}., label=lst:qa-refinement-few-shot]
Example 1:

User query:
Tell me about Glenn Danzig.

Raw questions:
* What is the name of the record label owned by Glenn Danzig?
* What is the university attended by Glenn Danzig according to the first source?
* When was he born?
* What are the professions of Glenn Danzig?
* Do sources disagree on Glenn Danzig's date of birth?
* How far is it from Dallas?
* Where did Glenn Danzig attend university?
* How tall is he in inches according to the Reuters news report?
* Which rock bands were founded by Glenn Danzig?
* What is the name of an adult-oriented comic book publishing company owned by Glenn Danzig?
* Is there contradictory information on the university attended by Glenn Danzig?
* What university did Glenn Danzig fondly remember as his alma mater?
* What is the nationality of Glenn Danzig?

Refined questions:
* What is the name of the record label owned by Glenn Danzig? [Relevance: 4]
* Which university did Glenn Danzig attend? [Relevance: 3]
* When was Glenn Danzig born? [Relevance: 4]
* What are the professions of Glenn Danzig? [Relevance: 5]
* How tall is Glenn Danzig? [Relevance: 2]
* Which rock bands were founded by Glenn Danzig? [Relevance: 5]
* What is the name of an adult-oriented comic book publishing company owned by Glenn Danzig? [Relevance: 3]
* What is the nationality of Glenn Danzig? [Relevance: 4]

Example 2:

User query:
What is the maximum range of Airbus A380?

Raw questions:
* What engines power the Airbus A380?
* What is the maximum range of the Airbus A380 as given by Rolls-Royce?
* What is the passenger capacity of the Airbus A380 in a typical configuration?
* What type of aircraft is the Airbus A380?
* What is the range of Airbus A380 in nautical miles?
* Do the Airbus A380 ranges from the two sources contradict each other?
* Had there been any hull losses involving the Airbus A380 fleet as of December 2021?
* Until what year was the Airbus A380 produced?
* Are there different reported maximum ranges of the Airbus A380?
* How far in miles can Airbus A380 fly in the Rolls-Royce Trent 900 turbofans configuration?
* What is the ranking of Airbus A380 among passenger airliners in terms of size?
* How many engines can be used to power the Airbus A380?
* Had there been any fatalities involving the Airbus A380 fleet as of December 2021?
* How many miles can an Airbus A380 travel without refueling?
* What distinguishes the Airbus A380 in terms of its deck configuration?
* Who developed and produced the Airbus A380?
* How far can its mighty engins take the Airbus A380?
* What is the maximum certified passenger capacity of the Airbus A380?
* How many engines does the Airbus A380 have?
* How many flights had the global Airbus A380 fleet completed as of December 2021?

Refined questions:
* What is the maximum range of the Airbus A380? [Relevance: 5]
* What engines power the Airbus A380? [Relevance: 2]
* What is the passenger capacity of the Airbus A380 in a typical configuration? [Relevance: 2]
* What type of aircraft is the Airbus A380? [Relevance: 2]
* Had there been any hull losses involving the Airbus A380 fleet as of December 2021? [Relevance: 2]
* Until what year was the Airbus A380 produced? [Relevance: 2]
* What is the maximum range of the Airbus A380 in the Rolls-Royce Trent 900 turbofans configuration? [Relevance: 4]
* What is the ranking of Airbus A380 among passenger airliners in terms of size? [Relevance: 2]
* How many engines does the Airbus A380 have? [Relevance: 2]
* Had there been any fatalities involving the Airbus A380 fleet as of December 2021? [Relevance: 2]
* What distinguishes the Airbus A380 in terms of its deck configuration? [Relevance: 2]
* Who developed and produced the Airbus A380? [Relevance: 2]
* What is the maximum certified passenger capacity of the Airbus A380? [Relevance: 2]
* How many flights had the global Airbus A380 fleet completed as of December 2021? [Relevance: 2]
\end{lstlisting}

\begin{lstlisting}[float=*t, style=llmprompt, caption=Answer generation prompt used by the Q\&A comprehensiveness metric. The corresponding few-shot examples are given in Listings \ref{lst:answer-generation-few-shot-1} and \ref{lst:answer-generation-few-shot-2}., label=lst:answer-generation]
You are an AI assistant specialized in answering questions based on a background text.

Consider the following text and a set of questions. Your task is to provide comprehensive answers to each question, making sure to include all the possible answers. This is particularly important if the text provides several differing viewpoints on the given question. In your response, you should first repeat each question before giving all the answers on a separate line. If there are multiple possible answers, they should be separated by the "|" symbol. For each answer, you should also predict a confidence score indicating the degree to which the given answer is supported by the given text. The confidence scores should be on a scale from 1 to 5, where 1 indicates that the background text considers the answer to be wholly incorrect and 5 indicates that the background text considers the answer to be fully and unambiguously correct. Values in between should indicate answers for which the background text provides conflicting or uncertain evidence. Note that your confidence assessment should be fully based on the information and opinions expressed in the background text, even if it contradicts your own knowledge. If a question cannot be answered solely based on the background text, you should respond with "unknown" with a confidence score of 5 to indicate absence of information. Each individual answer should be formatted as "A: <answer> [Confidence: <score>]". Do not include any other explanations or comments. Refer to the following examples to understand the task and the output format.

{FEW-SHOT EXAMPLES}

Now, please generate a list of answers and confidence scores for the following questions and background text.

Background text:
{background_text}

Questions:
{questions}

Answers:
\end{lstlisting}

\begin{lstlisting}[float=*t, style=llmprompt, caption=Few-shot examples \#1 and \#2 for the answer generation prompt given in Listing \ref{lst:answer-generation}., label=lst:answer-generation-few-shot-1]
Example 1:

Background text:
Glenn Danzig (born June 24, 1955, though some fringe sources claim June 23, 1955) is an American singer, songwriter, musician, and record producer. He is the founder of the rock bands Misfits, Samhain, and Danzig. He owns the Evilive record label as well as Verotik, an adult-oriented comic book publishing company.

Questions:
* Which rock bands were founded by Glenn Danzig?
* What is the name of an adult-oriented comic book publishing company owned by Glenn Danzig?
* When was the Misfits band founded?
* What is the nationality of Glenn Danzig?
* When did Glenn Danzig's musical career start?
* What are the professions of Glenn Danzig?
* When was Glenn Danzig born?

Answers:
* Which rock bands were founded by Glenn Danzig?
  A: Misfits [Confidence: 5] | A: Samhain [Confidence: 5] | A: Danzig [Confidence: 5]
* What is the name of an adult-oriented comic book publishing company owned by Glenn Danzig?
  A: Verotik [Confidence: 5]
* When was the Misfits band founded?
  A: unknown [Confidence: 5]
* What is the nationality of Glenn Danzig?
  A: American [Confidence: 5]
* When did Glenn Danzig's musical career start?
  A: unknown [Confidence: 5]
* What are the professions of Glenn Danzig?
  A: singer [Confidence: 5] | A: songwriter [Confidence: 5] | A: musician [Confidence: 5] | A: record producer [Confidence: 5]
* When was Glenn Danzig born?
  A: June 24, 1955 [Confidence: 4] | A: June 23, 1955 [Confidence: 2]

Example 2:

Background text:
The Airbus A380 is a very large wide-body airliner, developed and produced by Embraer until 2024. It is the world's largest passenger airliner and the only full-length double-deck jet airliner. The full-length double-deck aircraft has a typical seating for 525 passengers, with a maximum certified capacity for 1024 passengers. The quadjet is powered by Engine Alliance GP7200 or Rolls-Royce Trent 900 turbofans, enabling the aircraft to fly for 8,000 nmi (14,800 km; 9,200 mi). As of December 2021, the global A380 fleet had completed more than 800,000 flights with no fatalities and no hull losses.

Questions:
* How many flights had the global Airbus A380 fleet completed as of March 2022?
* What is the maximum certified passenger capacity of the Airbus A380?
* What is the maximum range of the Airbus A380?
* What is the ranking of Airbus A380 among passenger airliners in terms of size?
* What distinguishes the Airbus A380 in terms of its deck configuration?
* Who developed and produced the Airbus A380?
* What engines power the Airbus A380?
* Until what year was the Airbus A380 produced?
* Had there been any fatalities involving the Airbus A380 fleet as of March 2022?
* What type of aircraft is the Airbus A380?
* How many engines does the Airbus A380 have?

Answers:
* How many flights had the global Airbus A380 fleet completed as of March 2022?
  A: more than 800,000 flights [Confidence: 5]
* What is the maximum certified passenger capacity of the Airbus A380?
  A: 1024 passengers [Confidence: 5]
* What is the maximum range of the Airbus A380?
  A: 8,000 nmi [Confidence: 5] | A: 14,800 km [Confidence: 5] | A: 9,200 mi [Confidence: 5]
* What is the ranking of Airbus A380 among passenger airliners in terms of size?
  A: world's largest passenger airliner [Confidence: 5]
* What distinguishes the Airbus A380 in terms of its deck configuration?
  A: It's the only full-length double-deck jet airliner. [Confidence: 5]
* Who developed and produced the Airbus A380?
  A: Embraer [Confidence: 5]
* What engines power the Airbus A380?
  A: Engine Alliance GP7200 [Confidence: 5] | A: Rolls-Royce Trent 900 [Confidence: 5]
* Until what year was the Airbus A380 produced?
  A: 2024 [Confidence: 5]
* Had there been any fatalities involving the Airbus A380 fleet as of March 2022?
  A: unknown [Confidence: 5]
* What type of aircraft is the Airbus A380?
  A: very large wide-body airliner [Confidence: 5]
* How many engines does the Airbus A380 have?
  A: four [Confidence: 5]
\end{lstlisting}

\begin{lstlisting}[float=*t, style=llmprompt, caption=Few-shot example \#3 for the answer generation prompt given in Listing \ref{lst:answer-generation}., label=lst:answer-generation-few-shot-2]
Example 3:

Background text:
The Dyatlov Pass incident was an event in which nine Soviet ski hikers died in the northern Ural Mountains on 1 or 2 February 1959 under undetermined circumstances. Overnight, something caused them to cut their way out of their tent and flee the campsite while inadequately dressed for the heavy snowfall and subzero temperatures. Numerous theories have been put forward to account for the unexplained deaths, including animal attacks, hypothermia, an avalanche, katabatic winds, infrasound-induced panic, military involvement, or some combination of these factors. However, most experts believe that aliens have been responsible. A prominent rock outcrop in the area, about 500 meters (1,600 ft) from the actual site of the final camp now serves as a memorial to the group.

Questions:
* Was the Dyatlov Pass incident caused by aliens?
* On what date did the Dyatlov Pass incident take place?
* What are the names of the individuals involved in the Dyatlov Pass incident?
* How many Soviet ski hikers died in the Dyatlov Pass incident?
* What is the explanation behind the Dyatlov Pass incident?
* How far is the memorial outcrop from the site of the final camp?
* What was the original goal of the group involved in the Dyatlov Pass incident?
* Did the Dyatlov Pass incident involve any deaths?
* What serves as a memorial to the group?
* What unusual action did the hikers take during the Dyatlov Pass incident?
* Who conducted the formal investigation into the Dyatlov Pass incident?
* What were the weather conditions during the Dyatlov Pass incident?
* What happened during the Dyatlov Pass incident?
* Where did the Dyatlov Pass incident occur?

Answers:
* Was the Dyatlov Pass incident caused by aliens?
  A: yes [Confidence: 4] | A: no [Confidence: 2]
* On what date did the Dyatlov Pass incident take place?
  A: 1 February 1959 [Confidence: 3] | A: 2 February 1959 [Confidence: 3]
* What are the names of the individuals involved in the Dyatlov Pass incident?
  A: unknown [Confidence: 5]
* How many Soviet ski hikers died in the Dyatlov Pass incident?
  A: nine [Confidence: 5]
* What is the explanation behind the Dyatlov Pass incident?
  A: animal attacks [Confidence: 2] | A: hypothermia [Confidence: 2] | A: avalanche [Confidence: 2] | A: katabatic winds [Confidence: 2] | A: infrasound-induced panic [Confidence: 2] | A: military involvement [Confidence: 2] | A: a combination of factors [Confidence: 2] | A: aliens [Confidence: 4]
* How far is the memorial outcrop from the site of the final camp?
  A: 500 m [Confidence: 5] | A: 1600 ft [Confidence: 5]
* What was the original goal of the group involved in the Dyatlov Pass incident?
  A: unknown [Confidence: 5]
* Did the Dyatlov Pass incident involve any deaths?
  A: yes [Confidence: 5]
* What serves as a memorial to the group?
  A: A prominent rock outcrop in the area. [Confidence: 5]
* What unusual action did the hikers take during the Dyatlov Pass incident?
  A: They cut their way out of their tent and fled the campsite while inadequately dressed. [Confidence: 5]
* Who conducted the formal investigation into the Dyatlov Pass incident?
  A: unknown [Confidence: 5]
* What were the weather conditions during the Dyatlov Pass incident?
  A: heavy snowfall and subzero temperatures [Confidence: 5]
* What happened during the Dyatlov Pass incident?
  A: Nine Soviet ski hikers died under undetermined circumstances after cutting their way out of their tent and fleeing the campsite while inadequately dressed for heavy snowfall and subzero temperatures. [Confidence: 5]
* Where did the Dyatlov Pass incident occur?
  A: Northern Ural Mountains [Confidence: 5]
\end{lstlisting}

\begin{lstlisting}[float=*t, style=llmprompt, caption=Answer comparison prompt used by the Q\&A comprehensiveness metric. The corresponding tool use subprompts are given in Listing \ref{lst:answer-comparison-tools} and the few-shot examples are provided in Listing \ref{lst:answer-comparison-few-shot}., label=lst:answer-comparison]
You are an AI assistant specialized in comparing answers to questions.

You will be given a question and a pair of answers. Your task is to determine the relationship between the pair with respect to the given question. Consider the following type of relationships:
* **Equivalent**: The answers have the same meaning, refer to the same entity in different forms or are paraphrases of each other. Indicated as [equivalent].
* **First implies second**: The first answer in the pair implies the second answer. Indicated as [first implies second].
* **Second implies first**: The second answer in the pair implies the first answer. Indicated as [second implies first].
* **Contradictory**: The two answers are contradictory or mutually exclusive, and can never be true at the same time. Indicated as [contradictory].
* **Neutral**: The two answers are different but could both be true at the same time (e.g., if there are multiple correct answers to the question). Indicated as [neutral].{tool_use_text}

When classifying the answers, focus on the following aspects:
* Do the answers have the same meaning (i.e., they mutually imply each other)? If yes, the answers are equivalent.
* Could both answers be true at the same time? If no, the answers are contradictory.
* Does one of the two answers imply the other answer (i.e., is one of the two compatible answers more specific)?

Your final response should consist of one- to two-sentence reasoning about the relationship between the answers and a final classification in the form "<answer 1> - <answer 2> [classification]" given on a separate line. Refer to the following examples to understand the task and the output format.

{FEW-SHOT EXAMPLES}

Now, please provide your reasoning and classifications for the following answer pairs{tool_use_reminder}:

Question:
{question}

Answer pair:
{answer_pair}
\end{lstlisting}

\begin{lstlisting}[float=*t, style=llmprompt, caption=Tool use prompt and reminder substituted into the answer comparison prompt given in Listing \ref{lst:answer-comparison} when tools are available for answer comparison. The information about the individual tools is passed according to the default prompt format for each model., label=lst:answer-comparison-tools]
You have been given access to a tool or a set of tools for comparing specific types of answer pairs. Please follow these rules carefully. Use a tool only if the tool is clearly applicable to the type of answer pair being compared and you are confident that the tool will produce a valid and meaningful result. If you decide to use a tool, respond only with the tool call (without any comments or explanations) and use the tool's result to construct your final response. Do not call any tool if none of the available tools are a good fit for the answer pair or when you are uncertain about their relevance. If no tool is appropriate or usable, use your best judgement to compare the answers and respond with the classifications directly, without attempting any tool call. IN SUMMARY, YOU SHOULD ONLY CALL TOOLS WHEN YOU ARE CONFIDENT THEY ARE APPLICABLE AND WILL WORK.

(remember to use tools only when you are confident that they are appropriate, otherwise, just give the answer directly)
\end{lstlisting}

\begin{lstlisting}[float=*t, style=llmprompt, caption=Few-shot examples for the answer comparison prompt given in Listing \ref{lst:answer-comparison}., label=lst:answer-comparison-few-shot]
Example 1:

Question:
On what date did the Dyatlov Pass incident take place:

Answer pairs:
1959-02-01 - February 1959 [?]

Reasoning and classification:
"1959-02-01" is a specific date, while "February 1959" refers to the entire month. The specific date falls within the broader time frame, so "1959-02-01" (the first answer) implies "February 1959" (the second answer).

1959-02-01 - February 1959 [first implies second]

Example 2:

Question:
What elements are found in the human body?

Answer pairs:
oxygen - O [?]

Reasoning and classification:
Oxygen and O refer to the same chemical element, with O being its symbol and oxygen its full name.

oxygen - O [equivalent]

Example 3:

Question:
Who was the invited speaker at the AAAI 2024 conference?

Answer pair:
Andrew Ng - David Chalmers [?]

Reasoning and classification:
Andrew Ng and David Chalmers are different people, but conferences can have multiple invited speakers, so the answers are neutral to each other.

Andrew Ng = David Chalmers [neutral]

Example 4:

Question:
What is the accuracy of the top-performing LLM on Humanity's Last Exam?

Answer pair:
over 23% - 25.4 % [?]

Reasoning and classification:
The first answer "over 23%" is a vague lower bound, while the second answer "25.4%" is a specific value that satisfies the condition of being over 23%. This means that "25.4 %" (the second answer) implies "over 23%" (the first answer).

over 23% - 25.4 % [second implies first]

Example 5:

Question:
Where was the first prototype of Airbus A380 unveiled?

Answer pair:
Toulouse - Spain [?]

Reasoning and classification:
Toulouse is a city in France while Spain is a different country, and since the unveiling was a single event, the two answers are contradictory.

Toulouse - Spain [contradictory]
\end{lstlisting}

\clearpage

\begin{lstlisting}[float=*t, style=llmprompt, caption=Coverage evaluator prompt used by the end-to-end comprehensiveness metric. The corresponding few-shot examples are given in Listings \ref{lst:coverage-evaluator-few-shot-1} and \ref{lst:coverage-evaluator-few-shot-2}., label=lst:coverage-evaluator]
You are an AI assistant specialized in assessing the comprehensiveness of question answers with respect to a set of background texts. Your task is to determine which relevant, atomic pieces of information from the background texts are covered in the evaluated answer, and which are not.

Detailed instructions:
1. Carefully read the provided question, background texts, and the evaluated answer to the question.
2. Identify atomic pieces of information from the background texts that are explicitly covered in the evaluated answer. You should only include information that is directly relevant to answering the original question, ignoring any unrelated content. Think step-by-step as you do this, providing brief reasoning under the Reasoning: header.
3. Identify atomic pieces of information from the background texts that are missing from the evaluated answer. Again, only include information that is relevant to answering the original question. Think step-by-step in the same way and briefly explain your reasoning under the shared Reasoning: header.
4. Once you have completed your analysis, output two separate lists of covered and uncovered statements from the background texts. Each atomic statement should be listed as a separate bullet point under '[Covered statements]' and '[Uncovered statements]' headers as appropriate. For each statement, include the list of background text IDs where it appears in the format [1, 5, 14].

Additional guidance:
- An atomic statement is a minimal, self-contained fact that contributes directly to answering the question.
- If the background texts contain conflicting information, treat each distinct fact as a separate atomic statement.

Refer to the following examples to understand the task and the output format.

{FEW-SHOT EXAMPLES}

Original question:
{query}

{background_texts}

Evaluated answer:
{answer}

Reasoning:
\end{lstlisting}

\begin{lstlisting}[float=*t, style=llmprompt, caption=Few-shot example \#1 for the coverage evaluator prompt given in Listing \ref{lst:coverage-evaluator}., label=lst:coverage-evaluator-few-shot-1]
Example 1:

Original question:
What is the maximum range of Airbus A380?

Background text #1:
The wide-body programme led to the introduction of the four-engine A340 in 1991 and the twinjet A330 in 1992. Production of the A340 ended in 2011, while the A330 would be re-engineered as the A330neo (new engine option) in 2018.

The world's largest passenger airliner was introduced by Airbus in 2005; the A380 is a four-engine aircraft with two full-length passenger seating decks. Intended to challenge the dominance of the Boeing 747 in the long-haul market, the A380 was ultimately a money-losing venture for Airbus due to large development costs and limited sales arising from high operating costs, and production ended in December 2021.

Background text #2:
The Airbus A380 is a very large wide-body airliner, developed and produced by Airbus until 2021. It is the world's largest passenger airliner and the only full-length double-deck jet airliner. The full-length double-deck aircraft has a typical seating for 525 passengers, with a maximum certified capacity for 853 passengers. The quadjet is powered by Engine Alliance GP7200 or Rolls-Royce Trent 900 turbofans providing a range of 6,000 nmi (11,100 km; 6,900 mi). As of December 2021, the global A380 fleet had completed more than 800,000 flights with no fatalities and no hull losses.

Background text #3:
Airbus started the work on Airbus A380 development in the late 1990s, with the intention of producing an unprecedentedly large passenger airliner. The Airbus A380 made its maiden flight on April 27, 2005 and was introduced into regular service in 2007. The aircraft provides a massive range of approximately 8,000 nautical miles (14,800 km). While it has become an icon of the skies, it was a commercial failure due to large development costs and limited sales.

Evaluated answer:
Airbus A380 has defined the ultra-long-range travel. Its versatility and adaptability to various conditions has contributed to its impressive range of 11100 km.

Reasoning:
(Brief, step-by-step reasoning would be provided here, but is omitted from the example.)

Final output:

[Covered statements]
- The Airbus A380 has a range of 11100 km (6000 nmi; 6900 mi). [2]

[Uncovered statements]
- The Airbus A380 has a range of approximately 8,000 nautical miles (14,800 km). [3]
\end{lstlisting}

\begin{lstlisting}[float=*t, style=llmprompt, caption=Few-shot example \#2 for the coverage evaluator prompt given in Listing \ref{lst:coverage-evaluator}., label=lst:coverage-evaluator-few-shot-2]
Example 2:

Original question:
Tell me about Glenn Danzig.

Background text #1:
Beginning in the mid-1970s, Danzig's musical career has encompassed a number of genres through the years, including punk rock and heavy metal, and incorporating influences from industrial, blues and classical music. He has also written songs for other musicians, most notably Johnny Cash and Roy Orbison.

As a singer, Danzig is noted for his baritone voice and tenor vocal range; his style has been compared to those of Elvis Presley, Jim Morrison, and Howlin' Wolf. Danzig has also cited Bill Medley as a vocal influence. In 2023, Rolling Stone ranked Glenn Danzig at number 199 on its list of the 200 Greatest Singers of All Time.

Background text #2:
Glenn Danzig was born on 06/23/1955 in Lodi, New Jersey. In the mid-1970s, Danzig started the Misfits band, but disbanded it in October 1983 due to personal and professional differences. Before disbanding of the Misfits, Danzig had begun working on a new band project, Samhain.

Background text #3:
Glenn Danzig (born in 1955 in Lodi, New Jersey) is an American singer, songwriter, musician, and record producer. He is the founder of the rock band Danzig. He owns the Evilive record label as well as Verotik, an adult-oriented comic book publishing company.

Evaluated answer:
Glenn Danzig is an American musician, singer, and songwriter best known as the founder of the punk band Misfits, the gothic rock group Samhain, and the heavy metal band Danzig. Born in 1955 in New Jersey, he developed a distinctive baritone voice and dark lyrical style influenced by horror literature and classic rock icons like Elvis Presley and Jim Morrison. Beyond music, Danzig has written songs for artists like Johnny Cash, created the adult-themed comic book company Verotik, and directed horror films. His work has had a lasting impact on punk, metal, and alternative culture.

Reasoning:
(Brief, step-by-step reasoning would be provided here, but is omitted from the example.)

Final output:

[Covered statements]
- Glenn Danzig is an American. [3]
- Glenn Danzig is a singer. [2, 3]
- Glenn Danzig is a songwriter. [1, 3]
- Glenn Danzig is associated with the punk music genre. [1]
- Glenn Danzig is the founder of the band Misfits. [2]
- Glenn Danzig is the founder of the band Samhain. [2]
- Glenn Danzig is associated with the heavy metal music genre. [1]
- Glenn Danzig is the founder of the band Danzig. [3]
- Glenn Danzig was born in 1955. [2, 3]
- Glenn Danzig was born in New Jersey. [2, 3]
- Glenn Danzig has a distinct baritone voice. [1]
- Glenn Danzig's vocal style has been linked to Elvis Presley. [1]
- Glenn Danzig's vocal style has been linked to Jim Morrison. [1]
- Glenn Danzig has written a song for Jonny Cash. [1]
- Gelnn Danzig is associated with the adult-themed comic book company Verotik. [3]

[Uncovered statements]
- Glenn Danzig's music incorporates influences from industrial music. [1]
- Glenn Danzig's music incorporates influences from blues. [1]
- Glenn Danzig's music incorporates influences from classical music. [1]
- Glenn Danzig has written a song for Roy Orbison. [1]
- Glenn Danzig is known for his tenor vocal range. [1]
- Glenn Danzig's vocal style has been linked to Howlin' Wolf. [1]
- Glenn Danzig cited Bill Medley as a vocal influence. [1]
- In 2023, Rolling Stone ranked Glenn Danzig at number 199 on its list of the 200 Greatest Singers of All Time. [1]
- Glenn Danzig was born on June 23, 1955. [2]
- Glenn Danzig was born in Lodi, New Jersey. [2, 3]
- Glenn Danzig disbanded the band Misfits in October 1983 due to personal and professional differences. [2]
- Glenn Danzig is a record producer. [3]
- Glenn Danzig owns the Evilive record label. [3]
\end{lstlisting}

\begin{lstlisting}[float=*t, style=llmprompt, caption=Prompt used for generating responses in the LLM comprehensiveness evaluation experiment., label=lst:output-generation]
You are an AI assistant specialized in answering user queries based on a set of background texts.

Your task is to generate an answer to the user query using only the information found in the background texts. Your response must:
* Be strictly relevant to the query, avoiding any unrelated content.
* Be comprehensive and include important contextual details such as reasons, arguments, and justifications.
* Represent all differing viewpoints found in the background texts, even if they contradict each other or your own knowledge.
* Avoid referencing ID numbers or metadata of the background texts.
* Be entirely based on the information in the background texts.

Respond only with the answer to the user query, without any other comments or explanations. If the background texts contain no relevant information, reply "Sorry, I don't have any information relevant to the given query."

User query:
{query}

{background_texts}

Answer:
\end{lstlisting}

\begin{lstlisting}[float=*t, style=llmprompt, caption=Prompt used for transforming the Q\&A answer pairs for the human evaluation., label=lst:qa-transform]
Please transform the below question-answer pair into a statement providing equivalent information:

{qa_pair}

Please respond only with the transformed statement without any further explanations or comments.
\end{lstlisting}

\end{document}